\documentclass[lettersize,journal]{IEEEtran}
\usepackage{amsmath,amsfonts}
\usepackage{algorithmic}
\usepackage{array}
\usepackage[caption=false,font=scriptsize, labelfont=sf,textfont=sf]{subfig}
\usepackage{textcomp}
\usepackage{stfloats}
\usepackage{url}
\usepackage{verbatim}
\usepackage{graphicx}
\usepackage{booktabs}
\usepackage{adjustbox}
\usepackage[ruled,vlined]{algorithm2e}
\usepackage{hyperref}
\usepackage{multirow}
\usepackage{wrapfig}
\usepackage{xcolor} 
\usepackage{color}
\usepackage{soul}
\definecolor{myblue}{rgb}{0.0, 0.5, 0.7}
\def\BibTeX{{\rm B\kern-.05em{\sc i\kern-.025em b}\kern-.08em
    T\kern-.1667em\lower.7ex\hbox{E}\kern-.125emX}}
\begin{document}

\title{LiDAR-HMR: 3D Human Mesh Recovery from LiDAR}

\author{Bohao Fan, 
    Wenzhao Zheng,
    Jianjiang Feng, \IEEEmembership{Member,~IEEE}
    Jie Zhou, \IEEEmembership{Fellow,~IEEE}
    \IEEEcompsocitemizethanks{ 
        \IEEEcompsocthanksitem Corresponding author: Jianjiang Feng.
        \IEEEcompsocthanksitem Bohao Fan, Wenzhao Zheng, Jianjiang Feng, and Jie Zhou are with the Department of Automation, Tsinghua University.
        \IEEEcompsocthanksitem E-mail: \{fbh19\}@mails.tsinghua.edu.cn, \{jfeng, jzhou\}@tsinghua.edu.cn, wenzhao.zheng@outlook.com.
        \IEEEcompsocthanksitem This work was supported in part by the National Natural Science Foundation of China under Grant 62376132 and 62321005.
    }
    }

\maketitle

\begin{abstract}
Human mesh recovery (HMR) holds significant utility in many applications. Studying HMR involving various types of sensors is necessary, as it enables the acquisition of human meshes in diverse scenes. Unlike HMR based on RGB images, HMR based on LiDAR has received considerably less attention in previous works. The major challenge in estimating human poses and meshes from sparse point clouds lies in the sparsity, noise, and incompletion of LiDAR point clouds. To address these challenges, we propose a LiDAR-based 3D human mesh recovery algorithm, called LiDAR-HMR. This algorithm involves estimating a sparse representation of a human (3D human pose) and gradually reconstructing the body mesh. To better leverage the 3D structural information of point clouds, we propose a point-cloud-to-SMPL pipeline that uses the original point cloud features to guide the reconstruction. The experimental results on four publicly available datasets demonstrate the effectiveness of LiDAR-HMR. The codes are available at \url{https://github.com/soullessrobot/LiDAR-HMR}.
%  \keywords{3D human pose estimation \and Point cloud processing}
\end{abstract}

\begin{IEEEkeywords}
3D human pose estimation, human mesh recovery, point cloud processing
\end{IEEEkeywords}

\section{Introduction}
\label{sec:intro}
3D human pose estimation (HPE) and human mesh recovery (HMR) in unconstrained scenes have long been goals in computer vision. Compared with HPE, HMR contains extra information on the shape and details of the human body, which extends to a wider range of applications such as human-computer interaction and identity recognition. Mainstream 3D HMR or HPE methods focus primarily on RGB inputs \cite{li2021hybrik, choi2020pose2mesh, yoshiyasu2023deformable, lin2021mesh, lin2021end, hu2023personalized, zou2023human, kamel2021hybrid, liu2023fast, wu2019unsupervised} or multimodal fusion \cite{zheng2022multi, cong2023weakly, zanfir2023hum3dil, weng20233d, chen2024towards}, with sparse research dedicated to point cloud input. In particular, LiDAR point clouds are often utilized as auxiliaries for RGB images in multimodal settings. However, compared with RGB images, point clouds are more robust to illumination conditions and offer better privacy protection, thereby deserving more research. With the introduction of several point cloud-based datasets \cite{sun2020scalability, fan2023human, dai2023sloper4d} and 3D HPE or HMR algorithms such as LPFormer \cite{ye2024lpformer} and LiDARCap \cite{li2022lidarcap}, the feasibility of using LiDAR point clouds alone to perceive human meshes has been demonstrated.

\begin{figure}[th]
\centering
\subfloat[Incomplete.]{\includegraphics[width=0.15\textwidth]{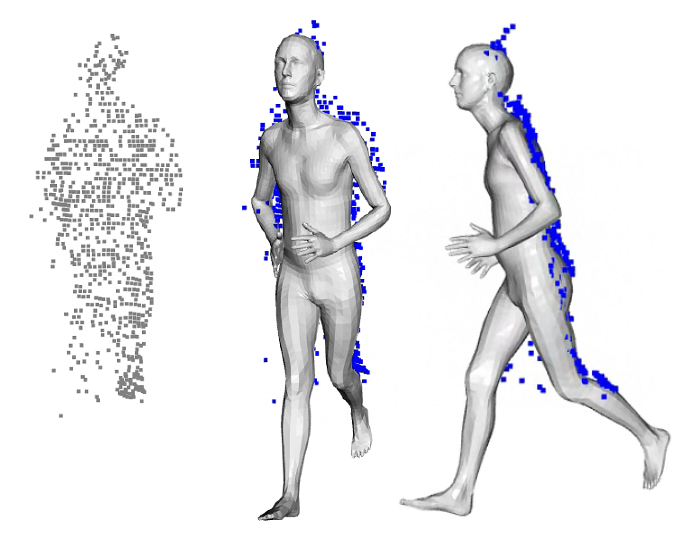}}%
\subfloat[Noise.]{\includegraphics[width=0.185\textwidth]{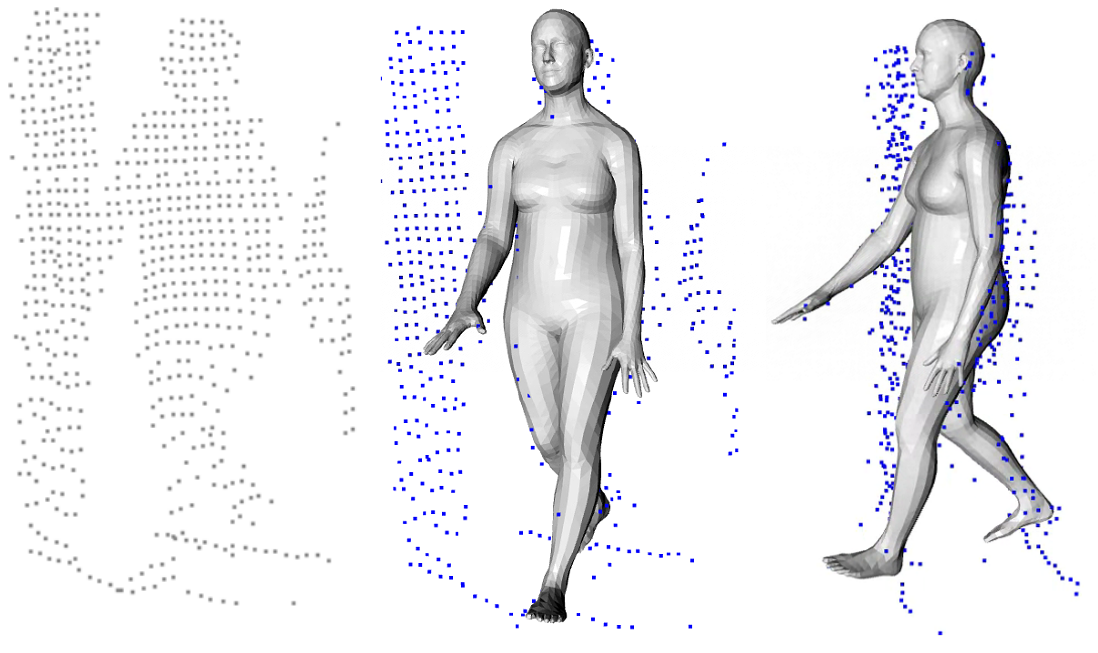}}%
\subfloat[Sparse.]{\includegraphics[width=0.163\textwidth]{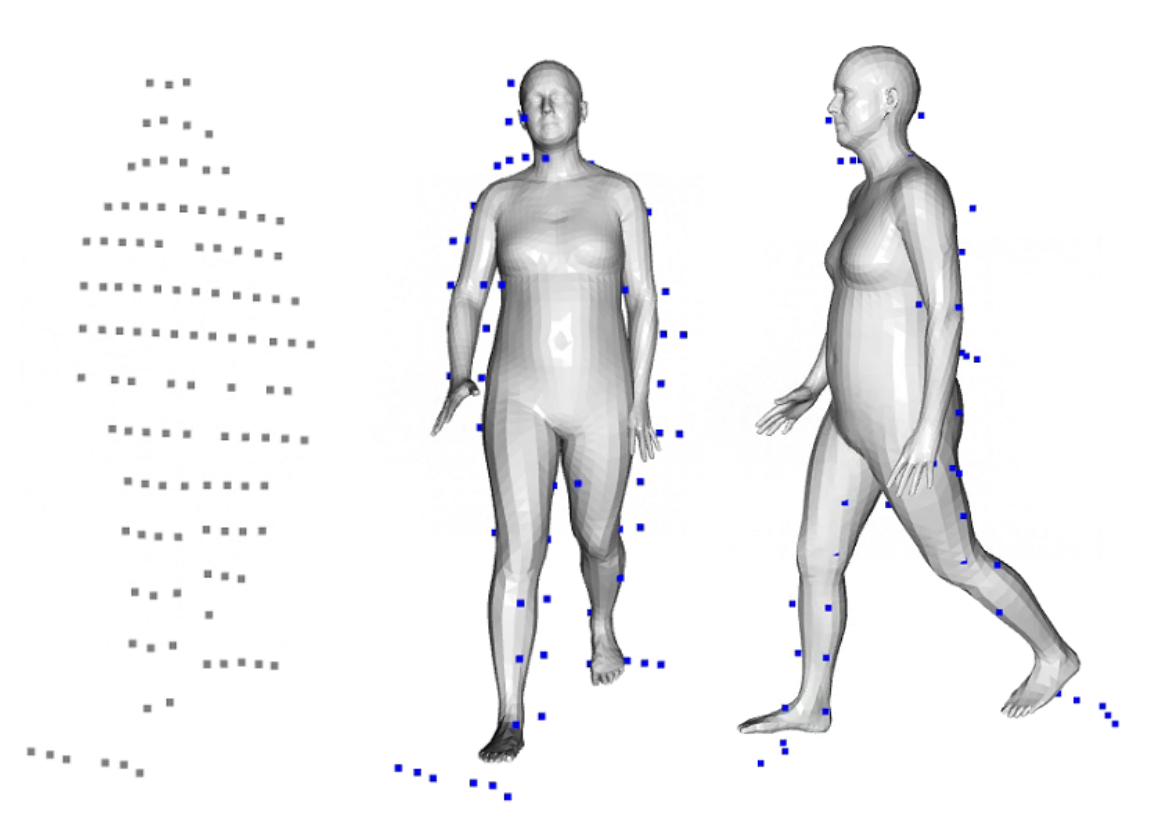}}%
\caption{Three challenges in 3D human mesh reconstruction from single frame sparse LiDAR point cloud. (Point clouds together with corresponding ground-truth meshes {(front view and side view)}.)}
\label{fig::difficult_and_pipeline}
\end{figure}

\begin{figure}
% {0.48\linewidth}
\centering
  \includegraphics[width=0.48\textwidth]{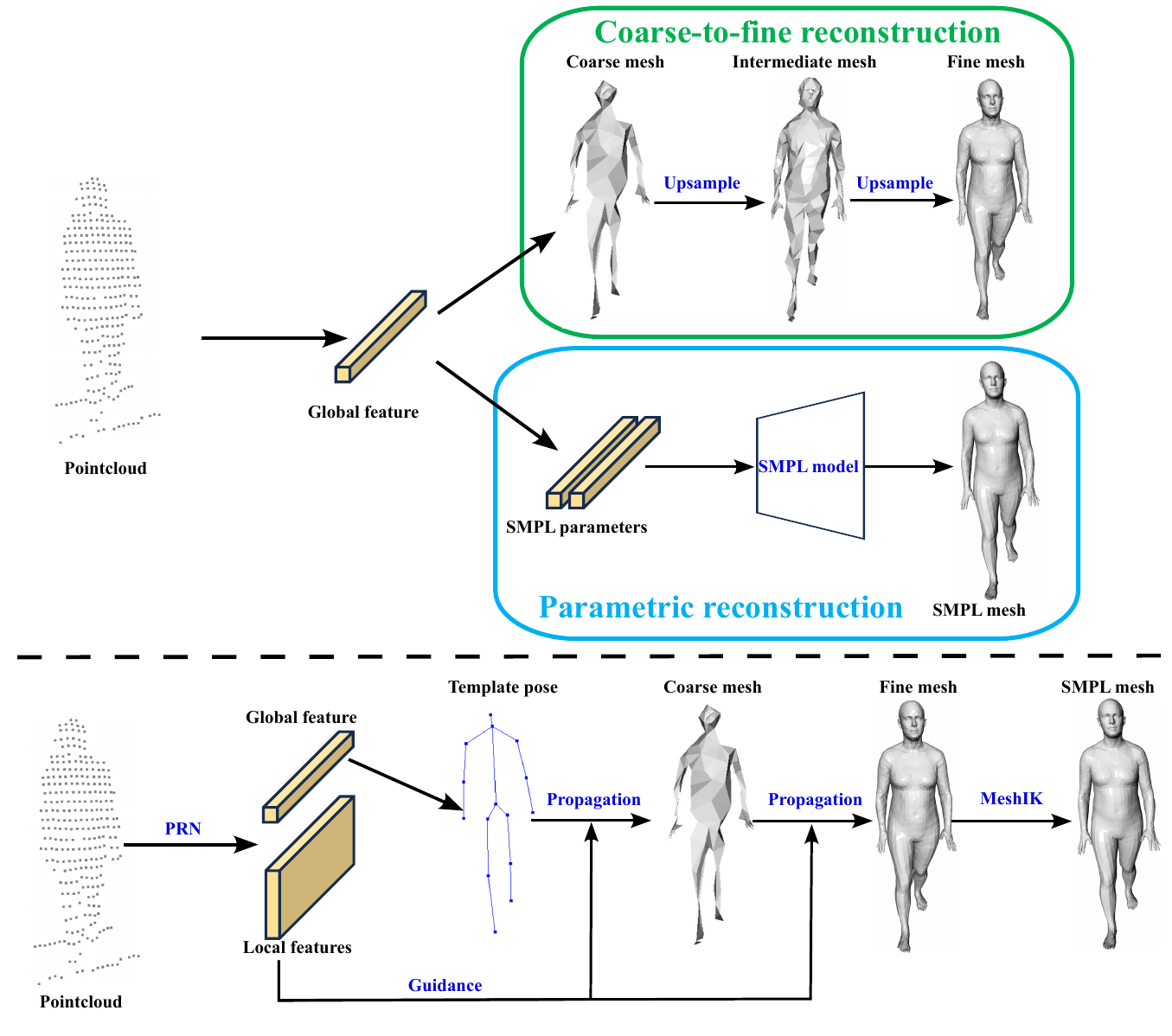}
  \caption{Top: Most 3D HMR methods from point cloud inputs utilize the sparse global feature extracted from the initial point cloud, which will cause information loss in local details. Down: The proposed point-cloud-to-SMPL pipeline for 3D HMR. Point cloud features are utilized to guide the coarse-to-fine process of human meshes to obtain better local details, and point features from coarse meshes are propagated rather than upsampled to obtain features in the fine meshes. Finally, a differentiable module, named MeshIK, can obtain SMPL parameters from fine meshes.}
    \label{fig:pipeline}
\vspace{-4mm}
\end{figure}

\begin{figure*}[t]
\centering
    \subfloat[Day scene in Waymo.]{\includegraphics[width=0.5338\textwidth]{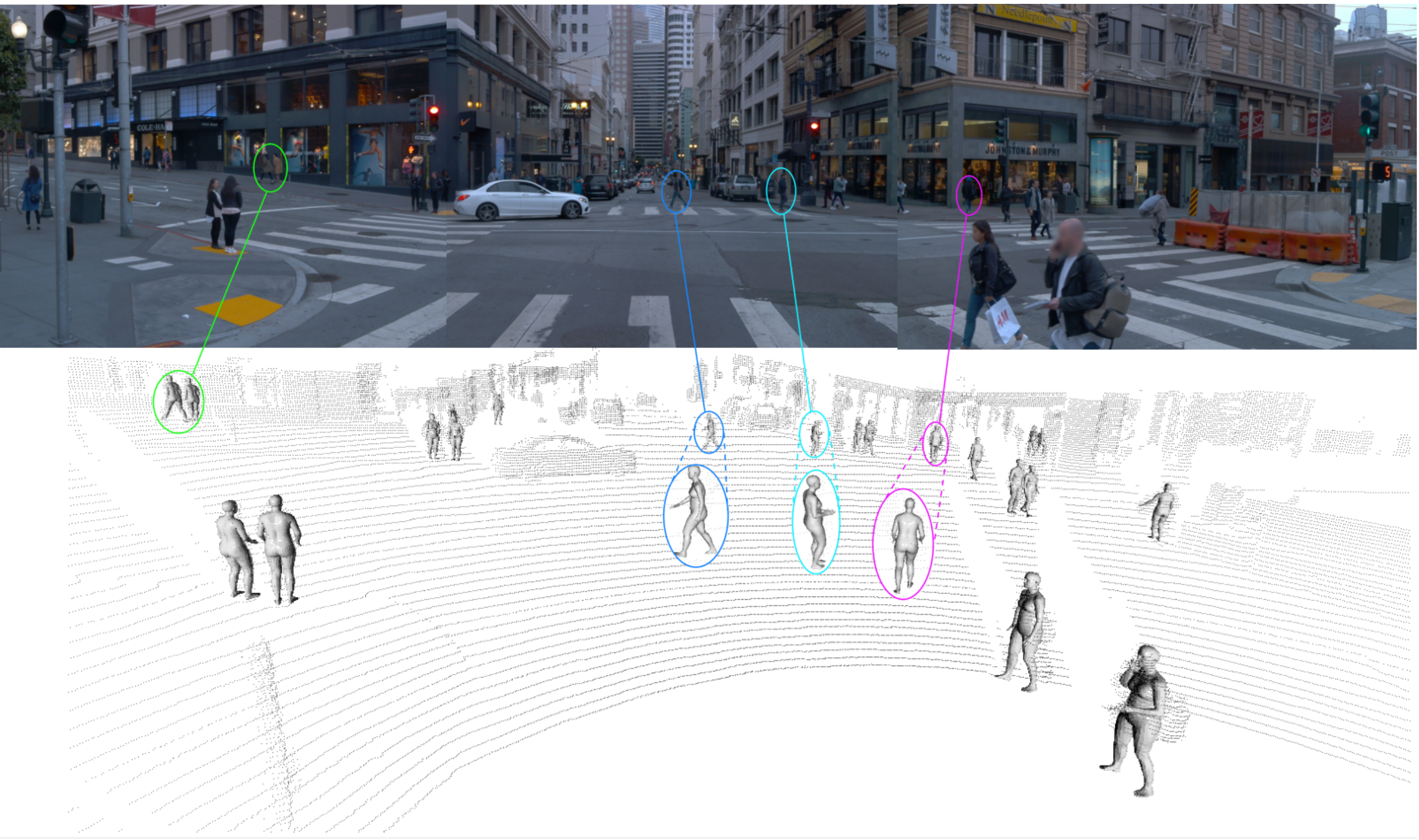}}
    \subfloat[Night scene in Waymo.]{\includegraphics[width=0.4752\textwidth]{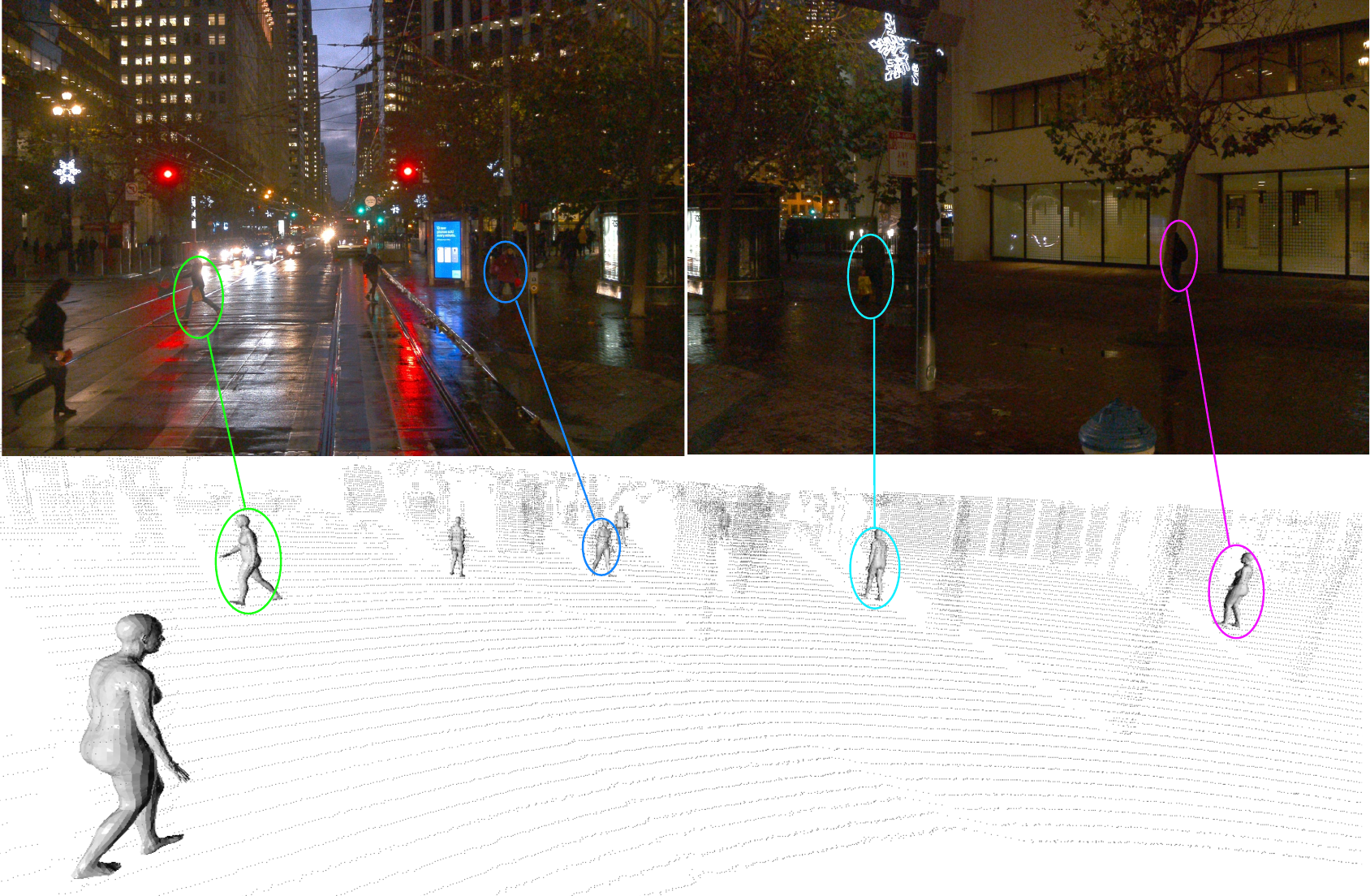}}
    \caption{Example results of LiDAR-HMR on multiperson scenes in the Waymo \cite{sun2020scalability} dataset. RGB images are not utilized in the algorithm but they are illustrated for better visualization. LiDAR-HMR can reconstruct accurate human meshes under different illumination conditions.}
    % , especially for the scene illustrated in (c): very faint illumination. }
    \label{fig:waymo_demo}
% \vspace{-3mm}
\end{figure*}

Existing 3D HMR methods based on single-frame point clouds \cite{liu2021votehmr, jiang2019skeleton} focus mainly on dense point clouds derived from depth maps. When captured by an RGB-D scanner in indoor settings, these point clouds are typically complete and dense, whereas the point clouds captured by LiDAR in outdoor scenes are sparse. Other methods \cite{li2022lidarcap, wang2020sequential} utilize sequential LiDAR point clouds for HMR. However, these methods must process a complete point cloud sequence, which means that the algorithm cannot output human meshes in real time, and cannot be trained using point clouds with only single frame labels, such as the Waymo dataset \cite{sun2020scalability}. Here, we consider a more practical and important setting: 3D HMR from single-frame LiDAR point clouds. This setting is analogous to image-based HMR, which is the basis of video-based HMR. This setting is quite challenging, as illustrated in Fig. \ref{fig::difficult_and_pipeline}, single-frame LiDAR point clouds have distinct characteristics: they are usually sparse, incomplete, self-occluding, and sometimes very noisy. 
As illustrated in Fig. \ref{fig:pipeline}, most point cloud-based HMR methods \cite{jiang2019skeleton,liu2021votehmr,li2021hybrik,li2022lidarcap} rely on neural networks to extract global features and regress SMPL parameters. However, the mapping from SMPL parameters to mesh vertices is inherently nonlinear. Direct regression of SMPL parameters from global features thus loses the correspondence between the point cloud and the human mesh vertices, which leads to a loss of information and diminishes its ability to fit data. This results in unsatisfactory HMR performance, especially under sparse point clouds. Specifically, \cite{wang2020sequential} employed a framework of coarse-to-fine reconstruction and regress mesh vertices, which are less affected by the above. They acquire a global feature representation from the input data, incrementally adding local details to reconstruct the human mesh. However, global feature extraction often leads to a loss of local details, and subsequent coarse-to-fine reconstruction relies on the human mesh priors to fill in the missing vertices, rather than referring to local surface details from the input point cloud, which results in the loss of specific local details. In addition, their outputs do not include explicit human pose semantics, but only mesh vertex coordinates, which limits their potential application.

To effectively reconstruct human poses from sparse point clouds, we aim to more comprehensively use the local details in the input point clouds. Consequently, we developed a point-cloud-to-SMPL dense reconstruction pipeline, which follows a sparse-to-dense reconstruction process. We use a pose regression network (PRN), which is a point cloud-based analysis network to extract template human 3D poses, perform pointwise segmentation and extract corresponding point cloud features. The extracted pointwise features can be directly utilized for mesh reconstruction in the subsequent mesh reconstruction network (MRN), thus constructing an end-to-end trainable network that combines point cloud analysis, pose estimation, and mesh reconstruction. Specifically, features between meshes of different resolutions are propagated on the basis of their relationships rather than being upsampled. Finally, we propose a differentiable MeshIK module to resolve the SMPL pose and shape parameters from the estimated 3D human mesh. The whole network is end-to-end trainable and we utilize end-to-end multitask (HPE and HMR) training for the entire network. As shown in Fig. \ref{fig:waymo_demo}, our LiDAR-HMR can handle various types of outdoor scenes. Compared with the RGB-based methods, it is not affected by low illumination , which can be an important advantage in certain applications. The experimental results on four public datasets demonstrate that LiDAR-HMR not only achieves the best performance in HMR but also achieves superior performance in HPE.

In summary, the contributions of this paper are as follows:

1. We propose a novel point cloud-based pipeline for human mesh recovery, which effectively incorporates local details from the point cloud into the reconstruction process to maximize the retention of local features.

2. We design a resolution-consistent feature propagation module that effectively inherits and constrains the feature relationships between meshes of varying resolutions throughout the reconstruction process from sparse to dense.

3. We emphasize and quantify the prevalent issue of local detail collapse in nonparametric mesh reconstruction methods, and we introduce a new module, MeshIK, to address this problem.

4. The proposed network, LiDAR-HMR, achieves state-of-the-art performance on several publicly available datasets.

\section{Related Work}
\label{related}

In recent years, human pose estimation (HPE) and human mesh recovery (HMR) have emerged as a prominent research topics, encompassing various areas of study, including 2D HPE \cite{cao2019openpose, xu2022vitpose}, 3D HPE \cite{haque2016towards, moon2018v2v, moon2019camera, zhen2020smap, xiong2019a2j, zanfir2023hum3dil, ye2024lpformer}, and human mesh recovery (HMR) \cite{kanazawa2018end, pavlakos2019expressive, kocabas2020vibe, lin2021mesh}. In this section, we focus on 3D HPE and HMR methods based on point cloud inputs. Given the relatively small number of 3D HMR works that focus on point cloud inputs, we also provide an overview of 3D HMR algorithms based on RGB images for reference in mesh recovery techniques.

\textbf{3D HPE from point clouds.} 3D HPE from depth images is a long-standing research topic \cite{knoop2006sensor, shotton2011real, girshick2011efficient, ganapathi2012real, ye2014real, helten2013personalization, moon2018v2v}.
However, depth image-based methods can only be used in indoor scenes. Recently, with the proposal of 3D human pose datasets for point cloud scenes \cite{dai2023sloper4d, li2022lidarcap, sun2020scalability}, several 3D pose estimation algorithms that are based on LiDAR point clouds have emerged. Zheng et al. \cite{zheng2022multi} first proposed a multimodal 3D HPE network to fuse RGB images and point clouds, using 2D labels as weak supervision for 3D pose estimation. In follow-up work, Weng et al. \cite{weng20233d} used simulation data to train a transformer-based pose estimation network and then proposed a symmetry loss to fine-tune with the actual LiDAR point cloud input. Ye et al. \cite{ye2024lpformer} proposed a multitask structure and used the object detection task as pretraining. They used a transformer as a point cloud encoder to regress human poses and achieved state-of-the-art effects on the Waymo \cite{sun2020scalability} dataset. These methods have successfully demonstrated the feasibility of estimating human body information from sparse LiDAR point clouds.

\textbf{3D HMR from point clouds.} 
Contemporary HMR algorithms for point clouds are mainly based on dense point clouds  \cite{jiang2019skeleton, liu2021votehmr} or sequential point clouds \cite{li2022lidarcap, wang2020sequential}. Specifically, Jiang et al. \cite{jiang2019skeleton} utilized a PointNet-based network to obtain point cloud features and used them to regress SMPL pose and shape parameters. Liu et al. \cite{liu2021votehmr} employed a voting mechanism to obtain sparse features of the human skeleton, and then used them to regress SMPL pose and shape parameters. Using point cloud video, Li et al. \cite{li2022lidarcap} proposed a GRU-based pipeline for sequential point cloud processing, and obtained sequential SMPL parameters via a spatial-temporal graph convolution network (ST-GCN). These algorithms fail to effectively synthesize the local details in the point clouds and are less effective for sparse point cloud inputs. Specifically, Wang et al. \cite{wang2020sequential} proposed a coarse-to-fine mechanism and reconstructed sequential human meshes via spatial-temporal mesh attention convolution. The reconstruction process mainly depends on the continuity of the sequence and the priori of the human body surface. Other works have attempted to address specific problems inherent in point clouds. Jang et al. \cite{jang2023dynamic} proposed a conditional variational autoencoder-based network to learn effective features from partial point clouds, with the simulated complete point clouds as guidance. And Zuo et al. \cite{zuo2021self} proposed a self-supervised HMR algorithm based on a Gaussian mixture model, which learns an effective conditioned posterior probability to remove noisy points. These methods are deeply based on simulation data; in real-world scenarios, it is difficult to obtain a sufficiently dense point cloud for full learning and modeling.

\textbf{3D HMR from RGB images.} 
3D HMR algorithms from RGB images can generally be divided into parameterized and nonparameterized methods. Most previous works \cite{kolotouros2019learning, kanazawa2018end, tung2017self, li2021hybrik} used parameterized human body models, such as SMPL \cite{loper2023smpl}, and focused on regressing parameters instead of the mesh itself. Given the pose and shape coefficients, SMPL is stable and practical for creating human meshes. However, as discussed in these papers \cite{omran2018neural, kolotouros2019convolutional, zhang2020learning, choi2020pose2mesh}, estimating the parameters directly from features is not easy because of the nonlinear mapping from the input data to the target parameters. Instead of regressing parametric coefficients, nonparametric methods \cite{choi2020pose2mesh, lin2021mesh, lin2021end} directly regress vertices from input data. In previous studies, graph convolutional neural networks (GCNNs) \cite{kolotouros2019convolutional, choi2020pose2mesh} were a popular option because they are able to model local interactions between adjacent vertices. However, they are less effective at capturing global features between vertices and body joints. To overcome this limitation, METRO \cite{lin2021end} proposed a set of transformer-based architectures to model the global characteristics of vertices. However, compared with GNN-based methods \cite{kolotouros2019convolutional, choi2020pose2mesh}, it is less convenient for modeling local interactions. In subsequent work, mesh graphormer \cite{lin2021mesh} used a combination of graph convolution and a transformer to obtain better results. 

In previous HMR methods based on coarse-to-fine mechanisms, sparse mesh representations are obtained from global features and lack sufficient 3D representation. During the reconstruction and fine-tuning process, the original features from the input data are not well encoded in the human body mesh reconstruction process. Specifically, the point cloud contains sufficient 3D occupancy clues and reliable human body surface details. We input the point cloud into the mesh reconstruction process to guide the local details of intermediate meshes.

\begin{figure*}[t]
\centering
    \subfloat[PRN structure.]{\includegraphics[width=0.35\textwidth]{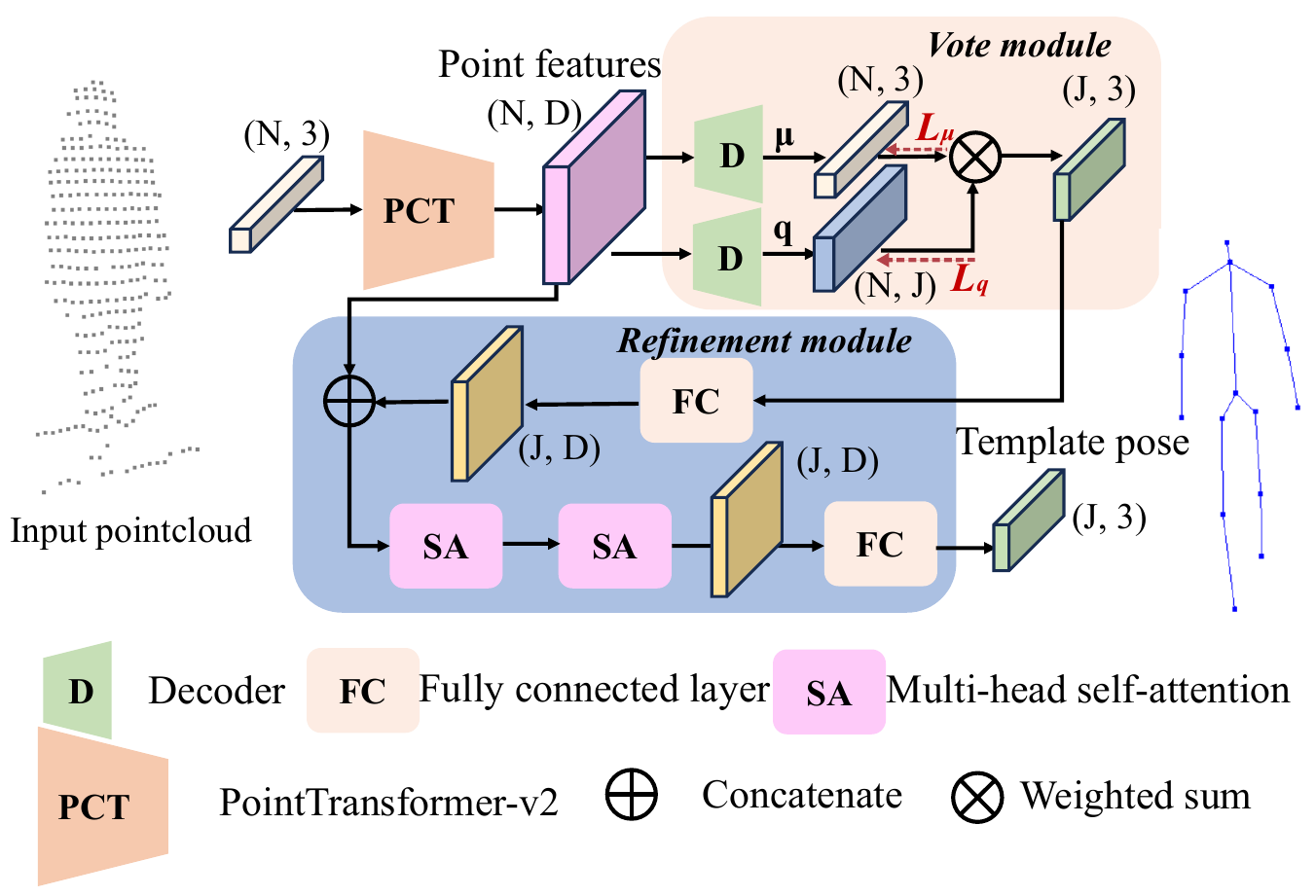}
    \label{fig:prn_structure}}
    \subfloat[MRN structure.]{\includegraphics[width=0.55\textwidth]{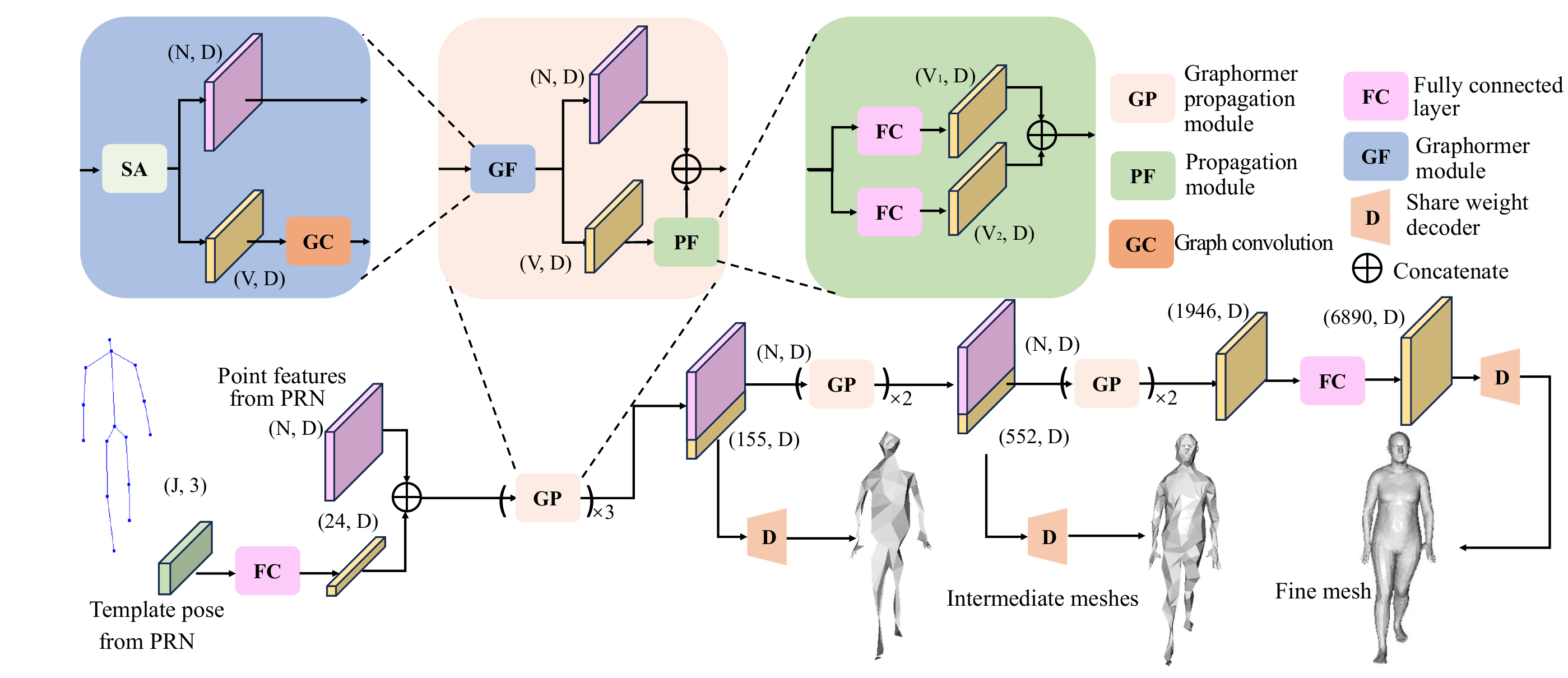}
    \label{fig:mrn_structure}}
    \caption{The overall structure of the proposed modules. (a) For the pose regression network (PRN), the input point clouds are encoded by PointTransformer-v2 and decoded into $q$ and $\mu$ to obtain a predicted human pose. The predicted pose is subsequently fed into two self-attention layers for refinement and completion. The red arrows indicate the loss functions for $q$ and $\mu$. The shape of the intermediate features is denoted as $(x,y)$. Specifically, $N$ denotes the number of input points, $J$ denotes the number of key points, and $D$ denotes the fixed feature dimension for attention.
    (b) The MRN receives the point cloud features, which estimates a template pose from the PRN and gradually reconstructs the complete human mesh. We utilize a point cloud-based graphormer \cite{lin2021mesh} for each intermediate resolution to introduce point cloud features during the reconstruction. Vertex features are inherited with a propagation module to model the parent-children relationship during the coarsening process. Finally, a fully connected layer is utilized to obtain the fine human mesh. The shape of the intermediate features is denoted as $(x,y)$. Specifically, $N$ denotes the number of input points, $V$ denotes the number of vertices, and $D$ denotes the fixed feature dimension for attention. }
\label{fig:network_structure}
\vspace{-3mm}
 \end{figure*}

\section{Method}
The proposed LiDAR-HMR network can be divided into three parts. As illustrated in Fig. \ref{fig:network_structure}, the PRN uses Point Transformer-v2 \cite{wu2022point} as a feature extractor to encode the point clouds, and it decodes the point features to reconstruct a template human pose. Given the template pose and per-point features, the MRN utilizes the point clouds to reconstruct the fine human mesh progressively. Finally, we propose MeshIK to obtain the SMPL pose and shape parameters from the estimated 3D mesh vertices.

\subsection{Pose Regression Network}
\textbf{Probabilistic modeling.}
With the input point cloud $\hat{P}$ of $n$ points $(p_0,p_1,...,p_{n-1})$, we model the problem of estimating 3D pose $J$ with $m$ key points $(j_0,...,j_{m-1})$ from point cloud $\hat{P}$ as follows:
\begin{equation}
        {\rm \max_{J}}  P(J|p_0,p_2,...,p_{n-1}).
\end{equation}
\indent Without the connection relationship between key points, we have:
\begin{equation}
\begin{aligned}
        P(J|p_0,p_2,...,p_{n-1})
        = \prod_{k=0}^{m-1}P(j_k|p_0,p_2,...,p_{n-1}).
\end{aligned}
\end{equation}
\indent With the hypothesis of the normal distribution and independent isodistribution, the probability distribution of $j_k$ can be described as follows:

\begin{equation}
\begin{aligned}
        P(j_k|p_0,p_2,...,p_{n-1})
        =&\prod_{i=0}^{n-1}P(j_k|p_i) \\
        =&\prod_{i=0}^{n-1} \frac{1}{\sqrt{2\pi} \sigma_{ik}} e^{{ -\frac{(j_k-\mu_{ik})^{2}}{{\sigma_{ik}}^{2}}} },\\
\end{aligned}
\end{equation}
where:
\begin{equation}
\begin{aligned}
        \mu_{ik} = p_i+d_{ik},
        p_i \in R_3,
        d_{ik} \in R_3.\\
\end{aligned}
\end{equation}
we assume that $d_i = d_{i0} = d_{i1} = ... = d_{im-1}$ is the unified pointwise offset, and that $\mu_{ik}$ and $\sigma_{ik}$ are parameters. We use a Gaussian distribution to approximate the joint posterior distribution as follows:
\begin{equation}
\begin{aligned}
        P(j_k|p_0,p_2,...,p_{n-1})\
        =&e^C\prod_{i=0}^{n-1} e^{{ -\frac{(j_k-\mu_{ik})^{2}}{{\sigma_{ik}}^{2}}}}\\
        =& e^C e^{\frac{(j_k-\hat{\mu}_k)^{2}}{{{\hat\sigma_k}}^{2}}},
\end{aligned}
\end{equation}

\begin{equation}
\begin{aligned}
        \frac{(j_k-\hat{\mu}_k)^{2}}{{\hat{\sigma_k}}^{2}}
        =-\sum_{i=0}^{n-1}\frac{(j_k-\mu_{ik})^{2}}{{\sigma_{ik}}^{2}},
\end{aligned}
\end{equation}

\begin{equation}
\begin{aligned}
C=-\frac{n}{2}ln(2\pi) - \sum_{i=0}^{n-1}ln(\sigma_{ik}).
\end{aligned}
\end{equation}

Letting $q_{ik} = \frac{1}{{\sigma_{ik}}^2}$, we have:
\begin{equation}
\begin{aligned}
        \hat{\mu}_k = \frac{\sum_{i=0}^{n-1}q_{ik}\mu_{ik}}{\sum_{i=0}^{n-1}q_{ik}}.
\end{aligned}
\end{equation}
\indent Letting $\sigma_{ik} \ge 1$ we obtain $q_{ik}\in(0,1]$, which describes the confidence of $p_i$ to vote for the position of keypoint $j_k$. The estimated model parameter $\hat{\mu} = (\hat{\mu}_0, \hat{\mu}_1, ..., \hat{\mu}_{m-1})$ denotes the estimated template pose $J_0$ with the maximized probability. 

\textbf{Estimate and complete human pose.}
The overall structure of the PRN is shown in Fig. \ref{fig:prn_structure}. We utilize PointTransformer-v2 \cite{wu2022point} as a per-point feature extractor and two decoders to regress $\mu_{ik}$ and $q_{ik}$. This process is similar to the ``vote'' operation in VoteNet \cite{ding2019votenet}; hence we name it the vote module. We further find that owing to the incomplete nature of the point cloud, one or more key points of the human body may not have been observed by the point cloud, leading to incomplete estimation of the template pose. To address this issue, we introduce a self-attention-based refinement module for completion and refinement, which consists of two self-attention layers. Notably, the refinement module does not have any point clouds or corresponding feature inputs. Instead, it relies on learned pose priors from the data.

\textbf{Loss.}
We use the L2 loss to constrain the estimated pose $J$ and ground truth human pose $J^{gt}$:
 \begin{equation}
        L_{J} =||J - J^{gt}||_2.
\end{equation}

To directly constrain the per-point features, we constrain the output of the vote module. Specifically, for each point $p_i$ in the input point cloud, we can obtain the ground truth values $\check{\mu_i}$ and $\check{q_i}$ from the ground truth pose. Then we use L2 loss and cross-entropy loss to constrain them:
\begin{equation}
        L_{\mu} =\sum_{i=1}^{N}||\mu_i - \check{\mu_i}||_2, 
        L_{q} =\sum_{i=1}^{N}CrossEntropy(Q_i, \check{Q_i}),
\end{equation}
where $\mu_i$ and $Q_i$ are the estimated values, $Q_i = (q_{i1}, ..., q_{iJ})$, and $\check{Q_i} = (\check{q_{i1}}, ..., \check{q_{iJ}})$, J is the number of key points. The overall loss function for the proposed PRN is as follows:
\begin{equation}
\label{eq_prn}
        L_{\mathrm{PRN}} = L_J + L_\mu + L_q,
\end{equation}
 
\begin{figure*}[t]
    % \vspace{-6mm}
    \includegraphics[width=1.0\textwidth]{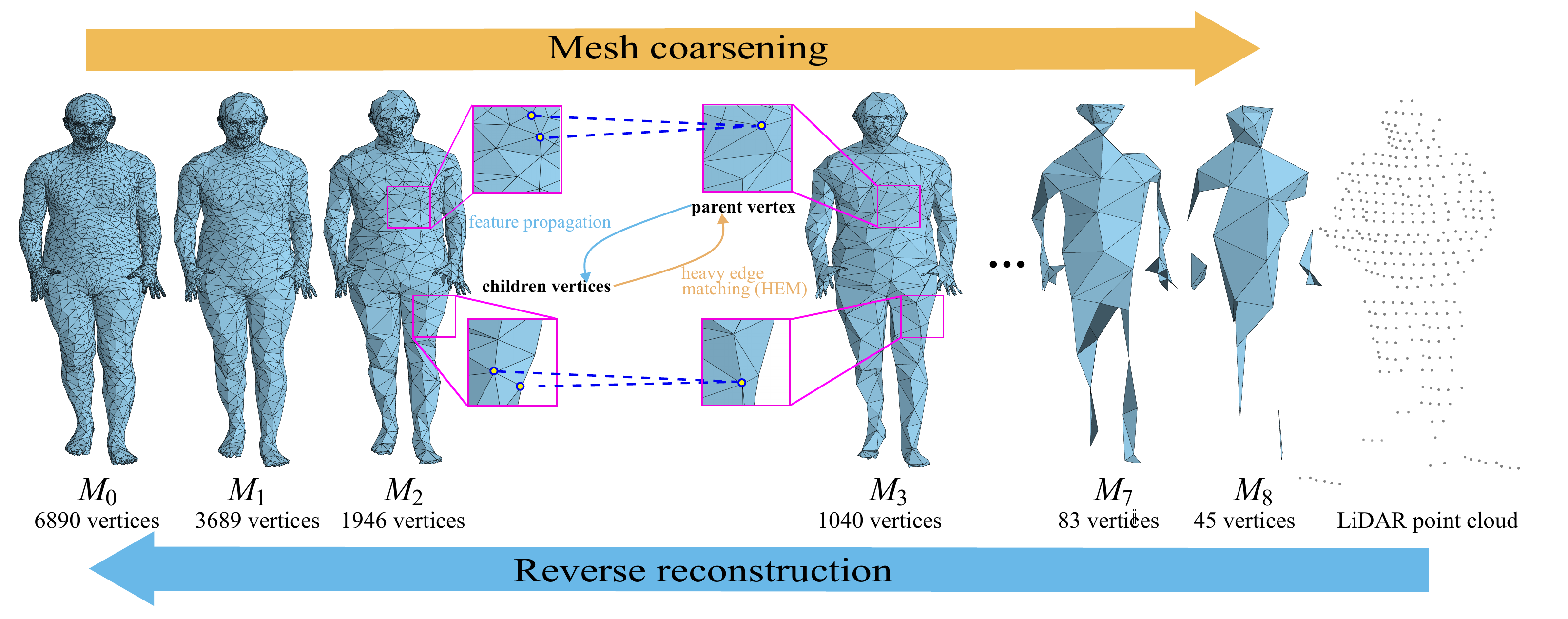}
    \vspace{-4mm}
    \caption{The mesh coarsening gradually generates multiple coarse graphs with heavy edge matching (HEM) following \cite{choi2020pose2mesh}. Specifically, we perform reverse reconstruction with the MRN, which utilizes the parent-children relationship during coarsening. The vertex features are propagated following the parent-children edge to generate a higher-resolution mesh.}
\label{fig:hem_structure}
\vspace{-4mm}
 \end{figure*}
 
\subsection{Mesh Reconstruction Network}
\label{sec:mrn}
As illustrated in Fig. \ref{fig:hem_structure}, we use heavy edge matching to downsample the complete human body mesh a total of 9 times to obtain a set of human skeletal structures at different resolutions. The sparsest skeleton corresponds to 24 vertices, whereas the densest mesh possesses 6890 vertices. This process constitutes the transition from sparse to dense. The MRN estimates the sparse vertices using the input template pose and point cloud features from PRN and gradually reconstructs the complete human body mesh. The overall structure of the MRN is illustrated in Fig. \ref{fig:mrn_structure}. 

\textbf{Graphormer for single resolution processing.}
The template pose is input into a fully connected layer that outputs sparse vertex features. We then utilize the graphormer propagation module which consists of a point cloud-based graphormer \cite{lin2021mesh} and a propagation module. The point cloud-based graphormer consists of a cascade of self-attention layers followed by a graph convolutional layer. It is utilized to process the relationships between the point cloud and the vertices of a mesh. The vertex and point cloud features are concatenated and input into a self-attention layer to introduce fine local point cloud features. The features of the mesh vertices after the self-attention module are then input into a graph convolutional layer to model the link relationships between vertices. Notably, the initial point cloud features are derived from the PRN, and the point cloud features are updated after the self-attention module.

\textbf{Propagation module for interresolution conduction.}
To obtain higher-resolution vertex features, Pose2Mesh \cite{choi2020pose2mesh} uses an upsampling approach, which, in this random way, does not consider the correspondence between vertices from different resolution meshes. We found that the correspondence and connecting relationships remain unchanged during the point cloud upsampling process. Specifically, as illustrated in Fig. \ref{fig:hem_structure}, each parent vertex has 1-2 child vertices at a finer resolution. We model the point features between different resolution meshes on the basis of this correspondence, assuming that the feature of a child vertex is obtained by a specific offset from its parent vertex feature. This is consistent with the inverse process of downsampling. We use two different fully connected layers to model these offsets between the relationships of the different resolution meshes, which is the propagation module.

For the first seven resolutions, we use one graphormer module and one propagation module for upsampling. For the last two resolutions with 3679 and 6890 vertices, owing to the large number of vertices, the use of a graphormer module results in an excessive computational burden. Therefore, we use two layers of fully connected layers to obtain the fine human mesh. Notably, we use a unified decoder to decode the vertex features, enabling each resolution of these mesh features to be decoded into the corresponding 3D coordinates. This ensures the consistency of the features learned by the MRN at different resolutions. Furthermore, we can supervise not only the final mesh output but also every intermediate resolution.

\textbf{Loss.}
The final output mesh loss $L_F$ is consistent with that in \cite{choi2020pose2mesh}, which consists of vertex coordinate loss, joint coordinate loss, surface normal loss, and surface edge loss. In addition, we also apply a mesh loss based on intermediate resolution. In intermediate resolution $i$ (from 1 to 8), L1 loss is utilized:
\begin{equation}
        L_{v\_i} =||M_i - \check{M_i}||_1,
\end{equation}
where $M_i$ is the estimated mesh and $\check{M_i}$ is the corresponding ground truth mesh. Then we can obtain the overall mesh loss for intermediate resolutions:
\begin{equation}
        L_{\mathrm{inter}} = \sum_{i=1}^{N}L_{v\_i}.
\end{equation}
Finally, the entire network is trained end-to-end, so losses in the PRN are also involved in training:

\begin{equation}
        L_{\mathrm{MRN}} = L_{\mathrm{PRN}} + L_F + L_{\mathrm{inter}}.
\end{equation}

\begin{figure*}[t]
    \includegraphics[width=1.0\textwidth]{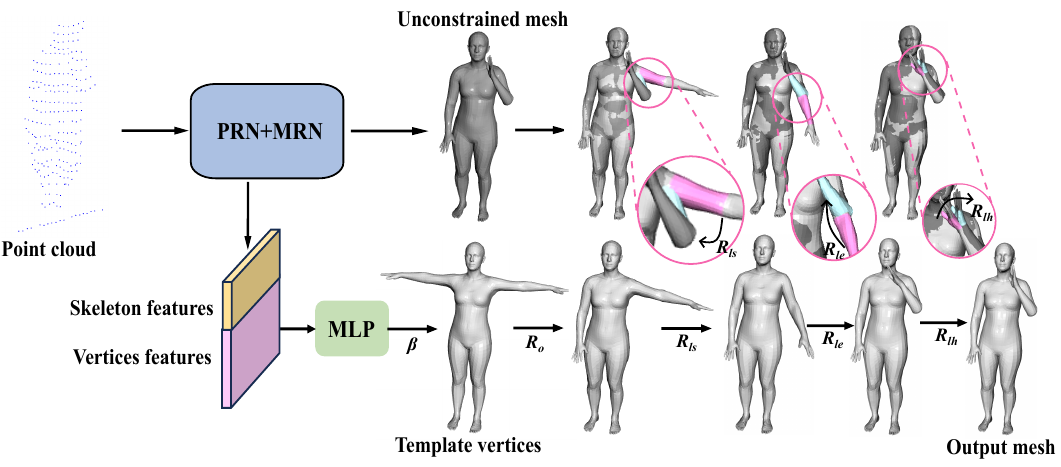}
    \vspace{-4mm}
    \caption{MeshIK reconstructs the SMPL pose and shape parameters from the unconstrained mesh estimated by MRN (gray mesh). Specifically, the template vertices (light mesh) are overlapped with the unconstrained vertices for better visualization. MeshIK follows the inference order of the SMPL model and gradually regresses each rotation matrix from the root. Here, we present an example of a left arm with key points: the left shoulder, left elbow, and left hand. Pink vertices indicate the selected local vertices for SVD in the template pose and the corresponding vertices in the unconstrained mesh are in light blue. ($R_{ls}$: rotation matrix of the left shoulder; $R_{le}$: rotation matrix of the left elbow; $R_{lh}$: rotation matrix of the left hand; $R_o$: rotation matrices of all other poses, which are obtained in the same way as the aforementioned three key points. )}
\label{fig:meshik}
\vspace{-3mm}
 \end{figure*}

\subsection{MeshIK: Inverse Kinematics from 3D Human Mesh}
Nonparametric algorithms such as MRN, Pose2Mesh \cite{choi2020pose2mesh} and MeshGraphormer \cite{lin2021mesh} output mesh vertex positions without constraints, which could cause unsatisfactory local details. Here, we propose a local point matching-based algorithm, MeshIK, to estimate the SMPL \cite{loper2023smpl} pose and shape parameters from the estimated mesh vertices. As illustrated in Fig. \ref{fig:meshik}, we first utilize an MLP to estimate the SMPL shape parameters of the human body. With the shape parameters, we can obtain the SMPL template vertices. Leveraging the cascading skeleton-driven orders of the SMPL model, we can resolve the related rotation matrix to ensure that the local surface point coordinates of the variables and the local surface point coordinates of the templates match as closely as possible. Fortunately, this can be accomplished using singular value decomposition (SVD). Specifically, for the predicted local mesh vertices $v = [v_0, ..., v_{N-1}]$ and local template vertices $t = [t_0, ..., t_{N-1}]$, the local rotation matrix $R_\theta$ to be solved is given by:
\begin{equation}
    R_\theta = \arg\underset{R}{\min}\sum_{i=0}^{N-1} w_i||v_i - Rt_i ||^2_2, \\
\end{equation}
where $w_i$ is the vertexwise weight calculated from the SMPL model parameters. Note that:
\begin{equation}
\label{meshik}
\begin{split}
   \min\sum_{i=0}^{N-1} w_i||v_i &- Rt_i ||^2_2 =  \min \sum_{i=0}^{N-1} ||w_iv_i - R(w_it_i) ||^2_2\\
   &\Leftrightarrow \min ||V_o - RT_o||^2_F \\
   &\Leftrightarrow \min \mathrm{trace}((V_o-RT_o)^T(V_o-RT_o))\\
   &\Leftrightarrow \max \mathrm{trace}(V_o^TRT_o) \\
   &\Leftrightarrow \max \mathrm{trace}(RT_oV_o^T), \\
\end{split}
\end{equation}

where 
\begin{equation}
\begin{split}
    V_o &= [w_0 v_0, w_1 v_1, ..., w_{N-1} v_{N-1}],\\
    T_o &= [w_0 t_0, w_1 t_1, ..., w_{N-1} t_{N-1}].
\end{split}
\end{equation}
$||\cdot||_F$ denotes the Frobenius norm, and $V_o$ and $T_o$ are weighted coordinate matrices. We can obtain the singular value decomposition of $T_oV_o^T$ as $V$ and $U$:
\begin{equation}
T_oV_o^T = UDV^T,
\end{equation}
where $D$ is a diagonal matrix and each element of $D$, $d_{ii} \geq 0$. Then:
\begin{equation}
\begin{split}
\max \mathrm{trace}(RT_oV_o^T) &\Leftrightarrow \max \mathrm{trace}(RUDV^T)\\
&\Leftrightarrow \max \mathrm{trace}(DV^TRU).\\
\end{split}
\end{equation}
Note that $M = V^TRU$ is an orthogonal matrix. Hence each element of $M$, $|m_{ij}| \leq 1$. Then:
\begin{equation}
    \mathrm{trace}(DV^TRU) = \mathrm{trace}(DM) = \sum_{i=1}^3 d_{ii}m_{ii}\leq{d_{ii}}.
\end{equation}
The trace is maximized when $m_{ii}=1, \forall 1\leq i \leq3$, where $M$ is the identity matrix.
Hence:
\begin{equation}
    V^TRU = I \Rightarrow R_\theta = VU^T.
\end{equation}

\textbf{Algorithms and pseudo code.} It is worth noting that owing to the rotation relationship of the human skeleton, the rotation matrix of the parent node will affect all its child nodes, hence the implementation varies with different poses. Specifically, the SMPL \cite{loper2023smpl} model inference process follows the human skeleton tree of SMPL joints, and the global rotation matrices $A_i$ utilized for the linear blend shape (LBS) of key point $i$ are obtained by multiplying all parent key points' local rotation matrices $R$. MeshIK determines $R_i$ incrementally according to the skeleton tree. A simplified version of the pseudocode is illustrated in Algorithm \ref{algo}, in which we omit the steps of matrix shape adjustment.

\begin{algorithm}
\SetAlgoLined
\KwIn{$M$, $T$, $W$, $Ind$, $R$, ${R_r}$, $Pa$, $A$}
\KwOut{$R$, $\theta$}
$R$[0] = ${R_r}$, $A[0] = R_r$\\
\For{$i \leftarrow 1$ \KwTo \rm{len}($Pa$) - 1}{
    ${M\_ = M[Ind[i]], T\_ = T[Ind[i]]}$\\
    ${W\_ = W[:, Ind[i]]}$\\
    ${WA_P = (W\_*A) @ T\_}$\\
    \# Remove the impact of other key points.\\
    ${P_0 = M\_ - A[Pa[i]] @ WA_P}$\\
    ${T_0 = W\_[i] * T\_}$\\
    ${S = T_0 @ P_0}$\\
    ${U, \_, V = \mathrm{SVD}(S)}$\\
    ${A[i] = V @ U, R[i] = A[Pa[i]]^{-1} @ A[i]}$
}
$\theta = {\mathrm{Rotmat\_to\_axis\_angle}(R)}$
\caption{MeshIK in a PyTorch-like style.}
\label{algo}
\end{algorithm}
% \vspace{-2mm}
Specifically, $M$ is the estimated mesh vertices with shape (6890, 3); $T$ is the template mesh vertices estimated from the shape parameters, $T$ possesses the same shape as $M$; $W$ is the LBS weights of the SMPL model with shape (24, 6890); $Ind$ is the list of selected vertices index with bool type and length 24; $R$ is the estimated local rotation matrix with shape (24, 3, 3), which is initialized with zero; ${R_r}$ is the root rotation matrix estimated by the same method from HybrIK \cite{li2021hybrik}; $Pa$ is the parent index with shape (24), and for example, $Pa[3] = 0$; $A$ is the LBS rotation matrix with shape (24, 3, 3), which is initialized with zero. $\theta$ is the estimated SMPL pose parameters with shape (72). $@$ indicates matrix multiplication and $*$ indicates elementwise multiplication. 

Specifically, during implementation, $M$ and $T$ are normalized to the corresponding local coordinates with the current key points $i$, and finally, the utilized matrices $\hat{A}$ for the LBS with shape (24, 4, 4) are as follows:
\begin{equation}
    \hat{A_i} = \begin{pmatrix}
    A_i & t_i \\
    0 & 1 \\
    \end{pmatrix},
\end{equation}
where $t_i$ is the local shift of the template key points which can be calculated by $A_i$ and the template skeleton length $l_t$. 

Finally, the local vertices indices $Ind$ are obtained for each joint by thresholding the LBS weights $W$. For each joint $i$ and vertex $j$, we filter vertices that satisfy $W_{ij} > 0.85$ as selected vertices, which produces $Ind$.

\textbf{Loss.}
With the input mesh of MeshIK $M_{\mathrm{inp}}$, output $M_{\mathrm{out}}$ and the corresponding ground truth $\check{M}$, the overall loss of MeshIK is defined as follows:
\begin{equation}
    L_{\mathrm{MIK}} = L_{\mathrm{MRN}} + ||M_{\mathrm{inp}} - M_{\mathrm{out}}||^2_2 + L_F(M_{\mathrm{out}}, \check{M}).
\end{equation}
The final output mesh loss $L_F$ is consistent with that in \cite{choi2020pose2mesh}.

Essentially, MeshIK establishes the relationship between human mesh vertices and SMPL pose parameters, and equates the task of learning the SMPL pose parameters with the regression mesh vertices, which greatly reduces the learning difficulty in the point cloud scenario.

\subsection{Implementation Details}
The number of input points is set to 1024. The entire LiDAR-HMR is end-to-end trainable. In the actual training process, we divide the training process into two steps to achieve better convergence. First, we pretrain the PRN with a batch size of 64 for 50 epochs with $L_{PRN}$ until convergence. Then, we perform end-to-end training of the PRN, MRN, and MeshIK together in a batch size of 8 for 50 epochs with $L_{MIK}$. The network parameters are updated via Adam \cite{kinga2015method}, and the learning rate is set to $5 \times 10^{-4}$. Specifically, the learning rate is reduced by a factor of 2 at the 10th and 20th epochs for the subsequent end-to-end training.

\section{Experiments}
\subsection{Datasets}
We conduct our experiments on four public datasets: Waymo \cite{sun2020scalability}, Human-M3 \cite{fan2023human}, SLOPER4D \cite{dai2023sloper4d} and LiDARHuman26M \cite{li2022lidarcap}.

The Waymo open dataset \cite{sun2020scalability} releases the human key point annotations on the v1.3.2 dataset, which contains LiDAR range images and associated camera images. We use v2.0 for
training and validation. It possesses 14 key points annotation for each object. There are 8125 annotated human key points in the training set and 1873 in the test set.

The Human-M3 dataset \cite{fan2023human} is a multimodal dataset that captures outdoor multiperson activities in four different scenes. It possesses 15 key point annotations for each object. There are 80103 annotated human key points in the training set and 8951 in the test set.

The SLOPER4D dataset \cite{dai2023sloper4d} is an outdoor human motion capture dataset that captures several objects on the basis of LiDAR point clouds and RGB images. Ground-truth meshes are obtained via motion capture devices. Overall, six motion fragments exist in the released part. As there is no manual assignment of training and test sets by the author, we selected a fragment as the test set. As a result, there are 24936 annotated human meshes in the training set and 8064 in the test set.

The LiDARHuman26M dataset \cite{li2022lidarcap} is a 
human motion capture dataset that captures several objects on the basis of LiDAR point clouds and RGB images. Ground-truth meshes are obtained via motion capture devices. There are approximately 160K annotated human meshes in the training set and 24K in the test set.

Owing to the lack of 3D mesh annotation in the Waymo and Human-M3 datasets, we used key point annotations and input point clouds to reconstruct the pseudolabel of the human mesh. This process is similar to that of Smplify-X \cite{pavlakos2019expressive}, named Smplify-Cloud.
For input point cloud $Q$ and ground-truth joint annotation $P = [p_1, ..., p_N]$, $N$ is the number of points. We estimate the pose parameter $\theta$ and shape parameter $\beta$ of the SMPL \cite{loper2023smpl} model and the corresponding root translation $r$. Specifically, pseudolabels are achieved by minimizing the loss function that constrains observations and priors:
\begin{equation}
    [r_p, \theta_p, \beta_p] = \mathop{\arg\min}\limits_{r, \beta, \theta}(L_{op}), %(r,\beta, \theta, P, Q)
\end{equation}
\begin{equation}
    P_M,M = S\_(r,\beta,\theta),
\end{equation}
\begin{equation}
    L_{op} = \lambda_J L_J(P_M, P)\\
    + \lambda_p L_p(M, \beta, \theta),
\end{equation}
\begin{equation}
  L_p(M, \beta, \theta) = L_{sp}(\beta) + L_{pp}(\theta) +  L_{cp}(M),
\end{equation}
where $S\_$ indicates SMPL model inference, $M$ indicates the 3D mesh vertices and $P_{M}$ indicates the 3D human pose joints processed by the SMPL model. $L_J$ is the l2-loss between the estimated 3D joints and the annotation joints. $L_p$ is the prior loss inherited from Smplify-X \cite{pavlakos2019expressive}. Specifically, $L_{sp}$ constrains the human shape, $L_{pp}$ constrains the human pose and angle, $L_{cp}$ is the collision punishment for the estimated mesh, and $\lambda_J$ and $\lambda_p$ are hyperparameters. The pseudolabels can be obtained via the L-BFGS optimization algorithm.

Specifically, to check the quality of the generated pseudolabels, we use Smplify-Cloud to generate pseudolabels on the SLOPER4D dataset and evaluate them using ground-truth annotation. The evaluation results are illustrated in Table \ref{table:table_sloper4d}, with metrics at a low level, so the quality of the pseudolabels is guaranteed.

For the Waymo-v2 and Human-M3 datasets, we used the annotated root joints (pelvis joints) of each pedestrian as the center and selected the point clouds contained within a cube of 2 m side length as the input to the network. When applied to large scene data, we expect that the 3D positions of pedestrians can be estimated by existing state-of-the-art pedestrian localization networks. Specifically, to simulate the estimation errors involved in pedestrian localization in practical applications, as well as random rotations, we apply random translations in the range of [-0.2 m, 0.2 m] and random rotations in the range of [-0.25$\pi$, 0.25$\pi$] during the training of the network.

\begin{table}[t]
\setlength{\tabcolsep}{6pt}
\centering
% \captionsetup{type=table}
\caption{Evaluation of Smplify-Cloud in the SLOPER4D dataset. The metrics include MPJPE (cm), MPVPE (cm), and MPERE.}
\begin{tabular}{ c c  c  c} 
\toprule
 SLOPER4D & MPJPE & MPVPE & MPERE \\ 
\midrule
Smplify-Cloud & 1.06 & 2.26 & 0.061\\
\bottomrule
\end{tabular}
\label{table:table_sloper4d}
\vspace{-4mm}
\end{table}

\subsection{Evaluation Metrics}
For HPE, we utilize MPJPE \cite{ionescu2013human3} and PA-MPJPE \cite{gower1975generalized, choi2020pose2mesh},  and the mean per vertex position error (MPVPE) for HMR. In addition, we propose a new metric, the mean per edge relative error (MPERE), for nonparametric method evaluation.
% Specifically, if the output meshes are a parametric representation, the edge lengths between vertices are fixed, but
For nonparametric methods, the position relationships between vertices are not fixed, and MPERE can represent the reconstruction quality of the local details.

\setlength{\arraycolsep}{2pt}
\begin{figure*}[t]
\scriptsize
\centering
\renewcommand{\arraystretch}{1.0}
\begin{tabular}{c c c c c c}
\setlength\tabcolsep{1.0pt}
\adjustbox{valign=c}{\includegraphics[width=0.12\linewidth]{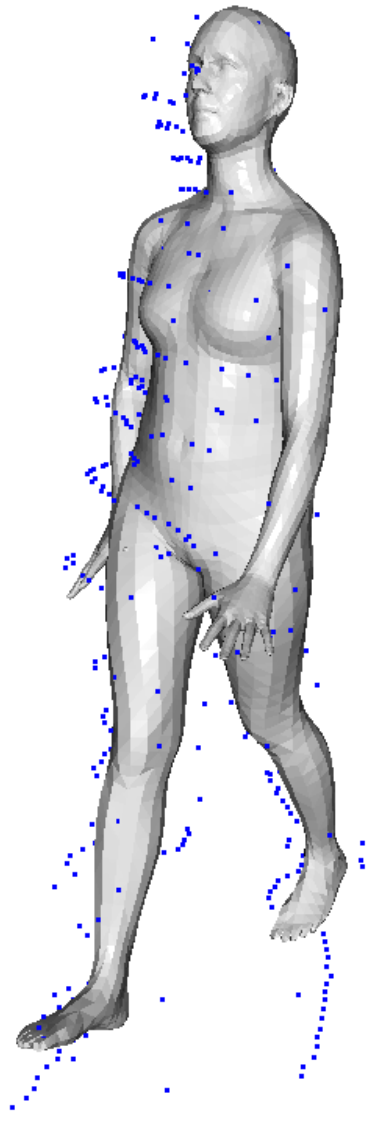}} & \adjustbox{valign=c}{\includegraphics[width=0.12\linewidth]{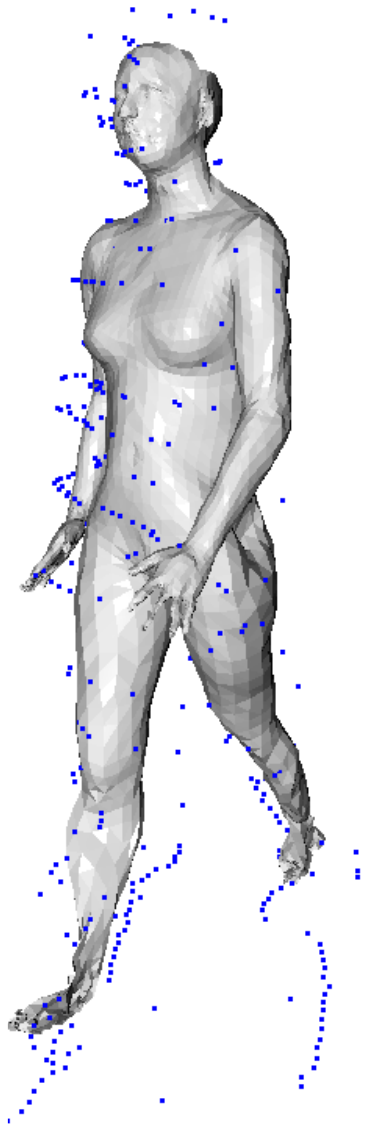}} & \adjustbox{valign=c}{\includegraphics[width=0.12\linewidth]{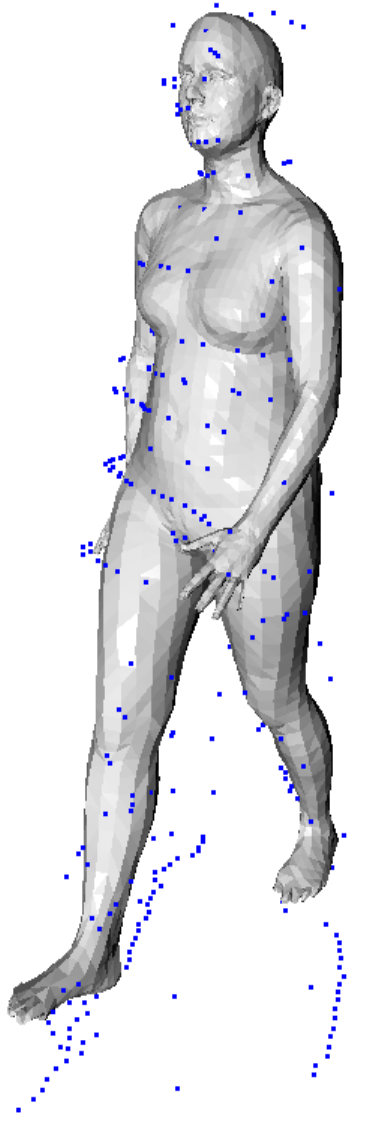}} & \adjustbox{valign=c}{\includegraphics[width=0.15\linewidth]{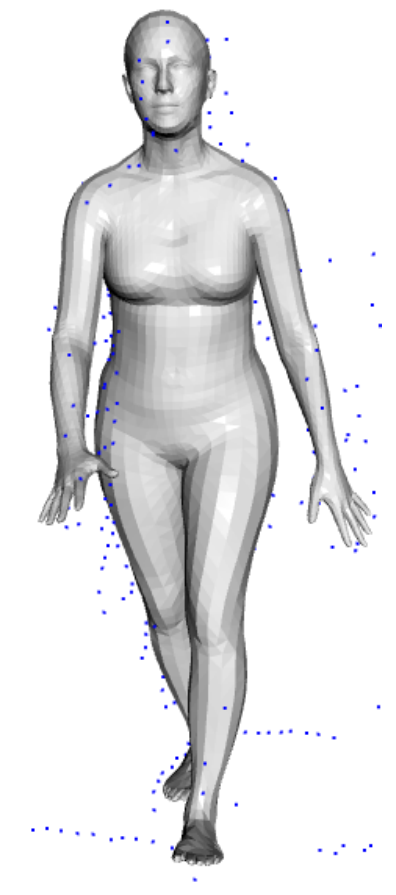}} & \adjustbox{valign=c}{\includegraphics[width=0.13\linewidth]{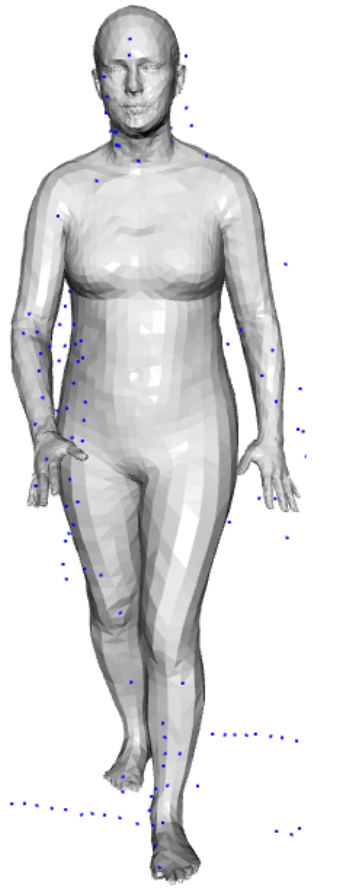}} & \adjustbox{valign=c}{\includegraphics[width=0.13\linewidth]{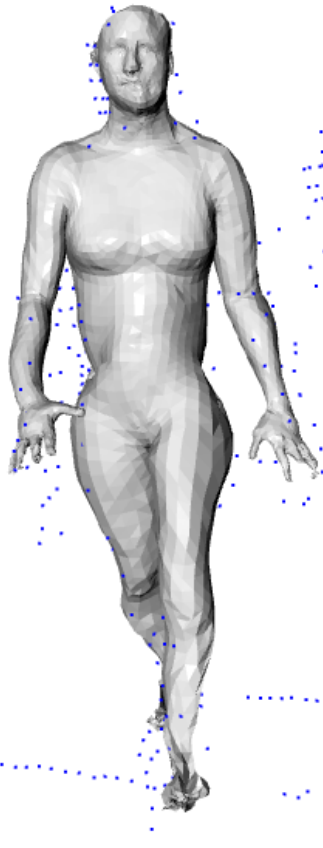}} \\
\multirow{2}{*}{Ground-truth} & VE = 5.37 cm & VE = 7.92 cm & \multirow{2}{*}{Ground-truth} & VE = 4.96 cm & VE = 4.50 cm\\
 & ERE = 0.135 & ERE = 0.109 & & ERE = 0.132 & ERE = 0.152
\end{tabular}
% \end{subfigure}
\centering
 % \vspace{-3mm}
\caption{Two examples of estimated human meshes, and point clouds are illustrated by the blue points. Meshes with low VE (MPVPE) may not have the best reconstruction quality. Therefore, we introduce the ERE (MPERE) to measure the reconstruction results from another perspective.}
% \vspace{-2mm}
\label{fig::diff_mpere}
\end{figure*}

For a predicted mesh $M$ with edge length set $(l_1,...,l_m)$, and ground-truth mesh $\check{M}$ with edge length set $(\check{l_1}, ..., \check{l_m})$, 
the MPERE is defined as follows:
\begin{equation}
        MPERE =\sum_{i=1}^{m} \frac{1}{m}\frac{||l_i - \check{l_i}||_1}{\check{l_i}},
\end{equation}
where $m$ is the number of edges in the mesh. 
MPERE measures the ratio of the length error to the ground-truth length of mesh edges and judges the reconstruction quality of short edges in dense parts more efficiently. As illustrated in Fig. \ref{fig::diff_mpere}, MPVPE may not be enough to measure the reconstruction quality. In this case, MPERE is needed as an additional measure.

In addition, we evaluated the computational cost during network training (GFLOPs) and the throughput during inference (frames per second, fps) of all the algorithms to compare the efficiency and lightness of the algorithms. Specifically, we use a GeForce RTX 3090 GPU as platform and record the network inference speed with a batch size of 1 in the test set of the Waymo-v2 dataset. For V2V-PoseNet and LPFormer, we follow the original setting where the volume cube size is set to 88*88*88.

\subsection{Comparison to the State-of-the-art}
For 3D HPE, we compare V2V-Posenet \cite{moon2018v2v} (V2V) and reimplement the LPFormer \cite{ye2024lpformer} (LP). For 3D HMR, we compare Pose2Mesh \cite{choi2020pose2mesh} (P2M), HybrIK \cite{li2021hybrik}, LiDARCap \cite{li2022lidarcap} (with single-frame input), reimplemented SAHSR \cite{jiang2019skeleton}, and VoteHMR \cite{liu2021votehmr}.
% our-implemented Mesh Graphormer \cite{lin2021mesh} with point cloud attention.

To evaluate the role of the PRN as a point cloud extraction backbone, we design and utilize the settings of LP+MRN+MeshIK for comparison. The LPFormer is pretrained, followed by end-to-end training of the entire network, as consistent with LiDAR-HMR. In addition, Pose2Mesh requires a well-estimated 3D human skeleton; hence, we utilize the results from the best 3D HPE model in the corresponding datasets as the input pose, called ``P2M''. For a fair comparison, we also utilize our proposed PRN (pretrained) to concatenate with MeshNet in the Pose2Mesh network for end-to-end training (the same as LiDAR-HMR), called ``PRN+P2M'' and ``PRN+P2M+MeshIK''. 

The quantitative evaluation results are illustrated in Table \ref{table:table_sota}. In particular, for \textbf{HPE}, the PRN achieves comparable results to those of state-of-the-art methods on different datasets but with significantly lower computational requirements. They rely on the voxelization of point clouds and 3D CNNs for extracting 3D features, which results in some computational redundancy and consumes more computational resources. Furthermore, the 3D features extracted by 3D CNNs do not provide significant assistance in the subsequent human mesh recovery. In addition, the reconstruction performance of ``V2V+P2M'' is weaker than that of the comparative group ``PRN+P2M''. This also demonstrates the effectiveness of our PRN network. The performance comparison between ``LP+MRN+MeshIK'' and ``LiDAR-HMR'' shows that although LPFormer can achieve good performance for pose estimation alone, when it is used as a part of the mesh reconstruction network, only the output of pose skeleton is not sufficient. Accordingly, the PRN can effectively extract effective features from the input point cloud, which is helpful for the subsequent mesh reconstruction.

\begin{table*}[t]
%\tiny
\setlength{\tabcolsep}{3.0pt}
\caption{Evaluation results on the SLOPER4D, Waymo, Human-M3 and LiDARHuman26M datasets. The metrics include JE (MPJPE, cm), VE (MPVPE, cm), and ERE (MPERE), together with the computation costs (GFLOPs) while training and the inference efficiency (fps) when batch size is set to 1. The proposed methods are shown in bold and the best values are shown in bold.}
\vspace{-4mm}
\begin{center}
% \resizebox{1\textwidth}{!}
{
\begin{tabular}{ c c c c c | c c c | c c c | c c c | c c} 
\toprule
 & & \multicolumn{3}{c}{SLOPER4D} & \multicolumn{3}{c}{Waymo} & \multicolumn{3}{c}{Human-M3} & \multicolumn{3}{c}{LiDARHuman26M} & \\
 & Model & JE & VE & ERE & JE & VE & ERE & JE & VE & ERE & JE & VE & ERE & GFLOPs & FPS \\ 
\midrule
\multirow{3}{*}{Pose} & V2V & \textbf{5.07} & - & - & 7.03 & - & - & 8.30 & - & - & 8.78 & - & - & 61.803  & 22.51 \\
& LP & 7.71 & - & - & \textbf{6.39} & - & - & \textbf{7.75} & - & - & \textbf{7.93} & - & -  & 62.197 & 23.92 \\
& \textbf{PRN} & 5.70 & - & - & 6.78 & - & - & 8.22 & - & - & 8.46 & - & - & \textbf{0.672} & 23.92 \\ 
\midrule
% \multirow{10}{*}{Mesh} & Graphormer & 7.71 & 9.23 & 1.689 & 8.05 
%   & 9.83 & 1.760 & 8.79 & 11.65 & 1.943 & & & & 9.15 \\
\multirow{10}{*}{Mesh} & HybrIK & 5.63 & 8.82 & 0.097 & 6.96 & 9.00 & 0.073 & 8.53 & 12.06 & 0.096 & 10.66 & 15.97 & 0.125 & 0.780 & 12.56 \\
& SAHSR & 7.26 & 8.12 & 0.085 & 9.66 & 11.68 & 0.163 & 10.55 & 13.15 & 0.291 & 10.70 & 13.34 & 0.111 & \textbf{0.427} & 24.39 \\
& VoteHMR & 5.46 & 6.09 & 0.079 & 9.76 & 12.08 & 0.194 & 10.58 & 12.54 & 0.189 & 13.14 & 16.22 & 0.211 & 0.663 & 21.81 \\ 
& LiDARCap & 9.79 & 10.56 & 0.06 & 11.10 & 13.19 & 0.061 & 10.12 & 11.51 & 0.032 & 9.35 & 12.52 & \textbf{0.036} & 0.522 & 77.32 \\
& V2V+P2M & 5.07 & 5.98 & 0.126 & 9.77 & 8.56 & 0.150 & 10.44 & 9.35 & 0.111 & 12.88 & 13.13 & 0.113 & 3.756 & 13.61\\ 
& PRN+P2M & 5.66 & 6.53 & 0.132 & 8.74 & 9.04 & 0.150 & 8.06 & 8.96 & 0.091 & \textbf{7.44} & 10.20 & 0.113 & 4.096 & 14.38 \\
& PRN+P2M+MeshIK & 6.36 & 6.27 & 0.042 & 7.65 & 9.24 & 0.067 & 8.12 & 9.02  & \textbf{0.034} & 7.54 & \textbf{10.18} & 0.089 & 4.180 & 7.55 \\
& LP+MRN+MeshIK & \textbf{4.31} & 5.46 & 0.063 & 6.86 & 8.82 & 0.067 & 7.84 & 9.22 & 0.047 & 7.57 & 10.39 & 0.045 & 63.854 & 7.52\\
& \textbf{PRN+MRN} & 4.76 & 5.19 & 0.094 & 7.01 & 8.24 & 0.119 & 7.94 & 8.95 & 0.088 & 7.63 & 10.39 & 0.079 & 2.225 & 12.35 \\ 
& \textbf{LiDAR-HMR} & 4.77 & \textbf{4.97} & \textbf{0.039} & \textbf{6.62} & \textbf{7.80} & \textbf{0.062} & \textbf{7.60} & \textbf{8.64} & 0.044 & 7.62 & 10.25 & 0.075 & 2.309 & 7.45\\ 
  % PRN+MRN+MeshIK
\bottomrule
\end{tabular}
}
\end{center}
\label{table:table_sota}
\vspace{-4mm}
\end{table*}

\begin{figure*}[t]
\scriptsize
\centering
% \vspace{-3pt}
%\scriptsize
 % default value: 6pt
\setlength{\arraycolsep}{0pt}
\renewcommand{\arraystretch}{1}
\resizebox{1\textwidth}{!}{
\begin{tabular}{c c c c c c c c c c c}
\setlength\tabcolsep{1.0pt}
Point cloud & GT & GT\_V & LMR & PM & P2M & PP & PPM & HybrIK \cite{li2021hybrik} & LiDARCap \cite{li2022lidarcap} & VoteHMR \cite{liu2021votehmr} \\
\adjustbox{valign=c}{\includegraphics[width=0.08\linewidth]{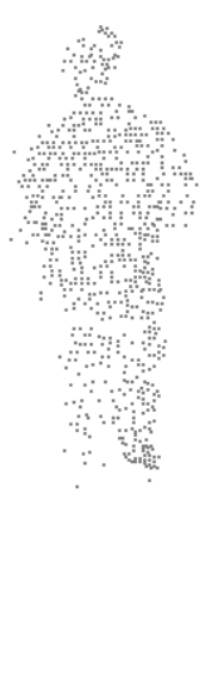}} & \adjustbox{valign=c}{\includegraphics[width=0.075\linewidth]{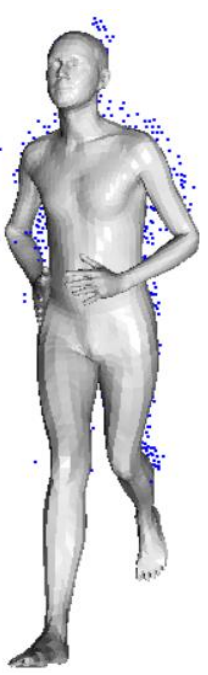}} & \adjustbox{valign=c}{\includegraphics[width=0.13\linewidth]{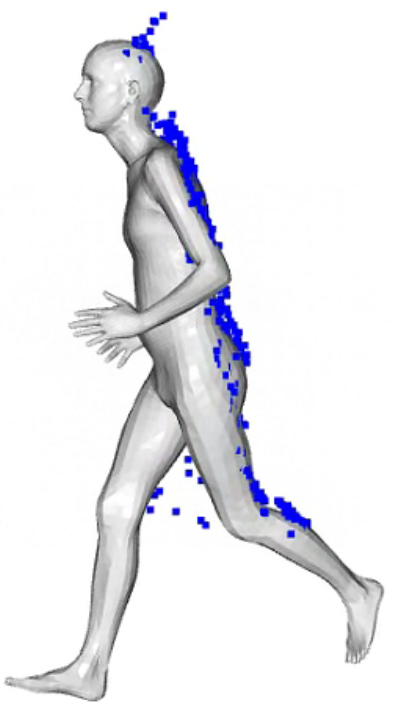}} & \adjustbox{valign=c}{\includegraphics[width=0.082\linewidth]{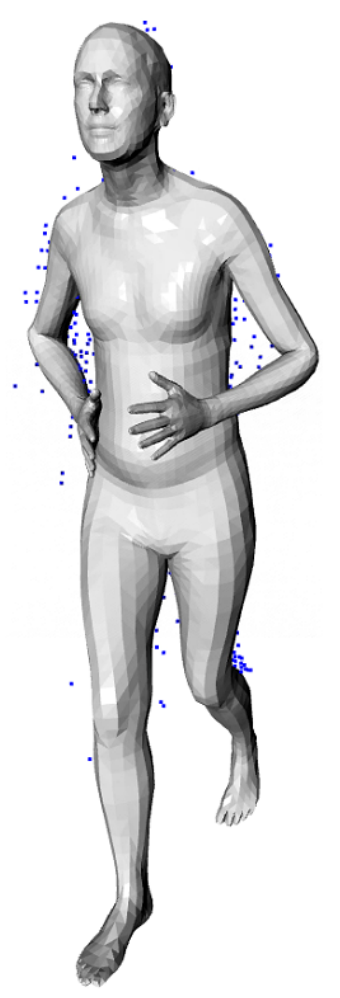}} & \adjustbox{valign=c}{\includegraphics[width=0.08\linewidth]{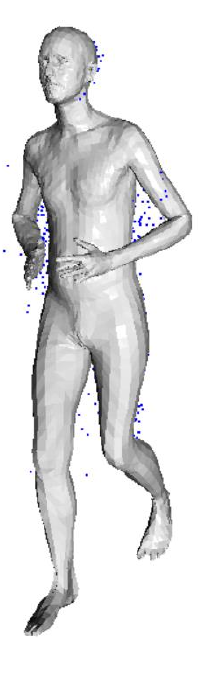}} & \adjustbox{valign=c}{\includegraphics[width=0.08\linewidth]{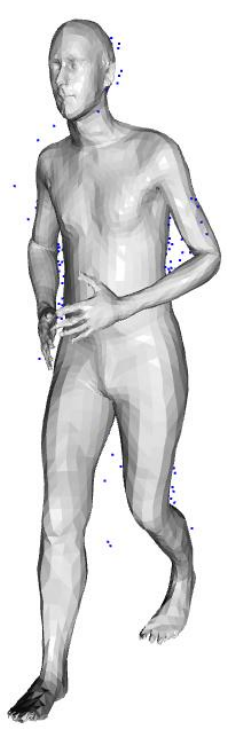}} & \adjustbox{valign=c}{\includegraphics[width=0.08\linewidth]{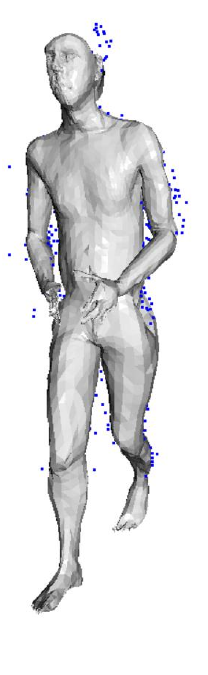}} & \adjustbox{valign=c}{\includegraphics[width=0.08\linewidth]{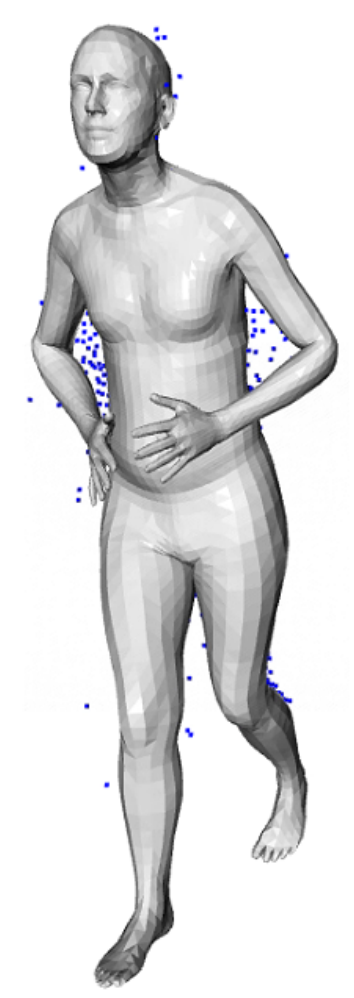}} & \adjustbox{valign=c}{\includegraphics[width=0.08\linewidth]{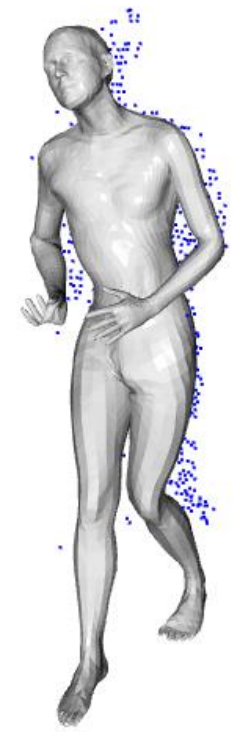}} & \adjustbox{valign=c}{\includegraphics[width=0.08\linewidth]{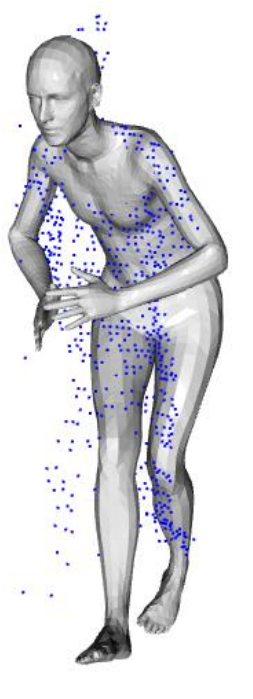}} & \adjustbox{valign=c}{\includegraphics[width=0.08\linewidth]{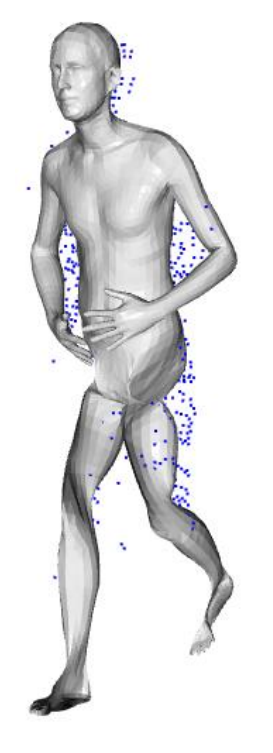}}
\\
\adjustbox{valign=c}{\includegraphics[width=0.09\linewidth]{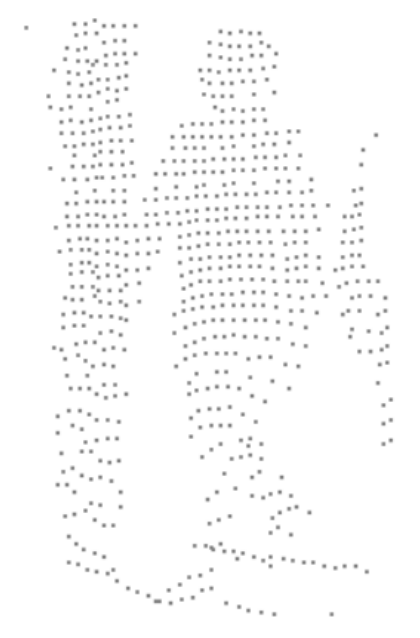}} & \adjustbox{valign=c}{\includegraphics[width=0.09\linewidth]{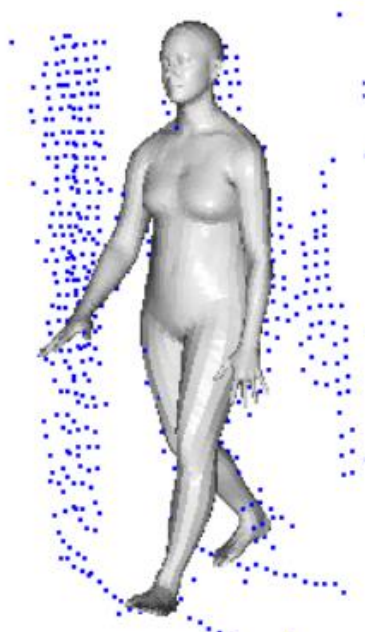}} & \adjustbox{valign=c}{\includegraphics[width=0.09\linewidth]{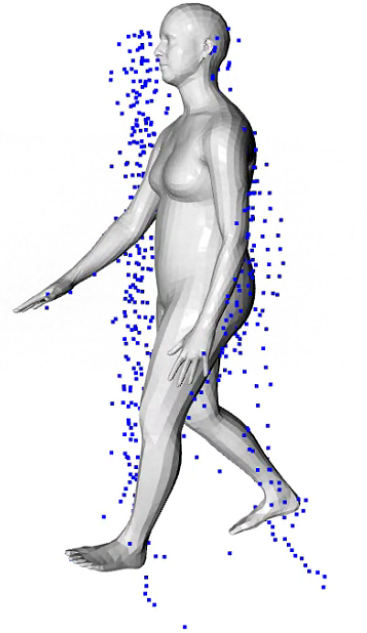}} & \adjustbox{valign=c}{\includegraphics[width=0.09\linewidth]{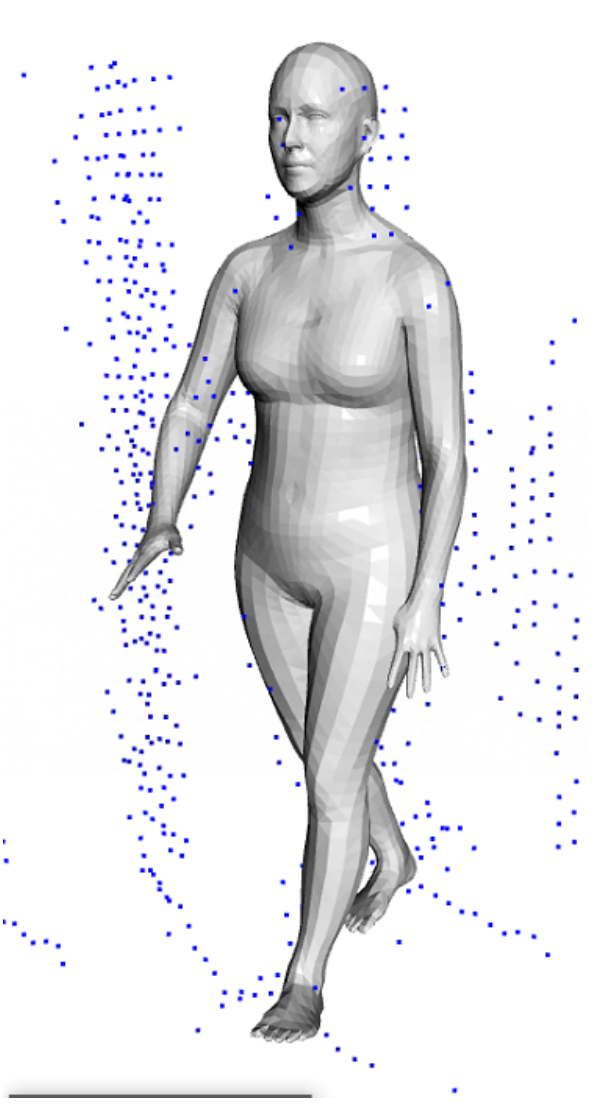}} & \adjustbox{valign=c}{\includegraphics[width=0.09\linewidth]{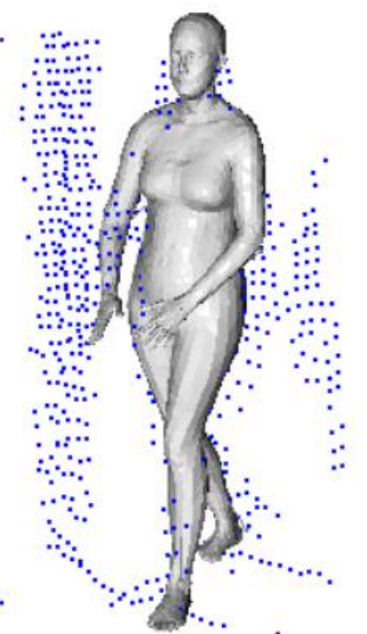}} & \adjustbox{valign=c}{\includegraphics[width=0.09\linewidth]{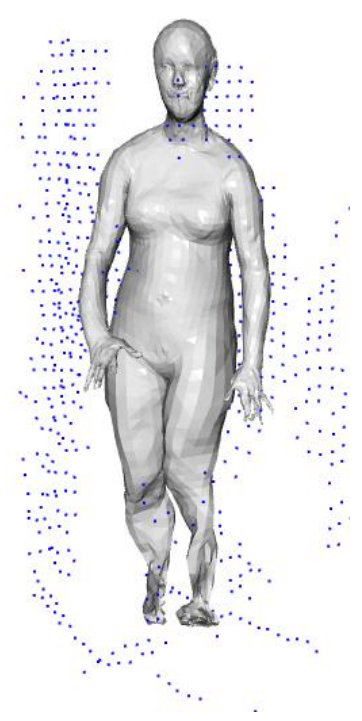}} & \adjustbox{valign=c}{\includegraphics[width=0.09\linewidth]{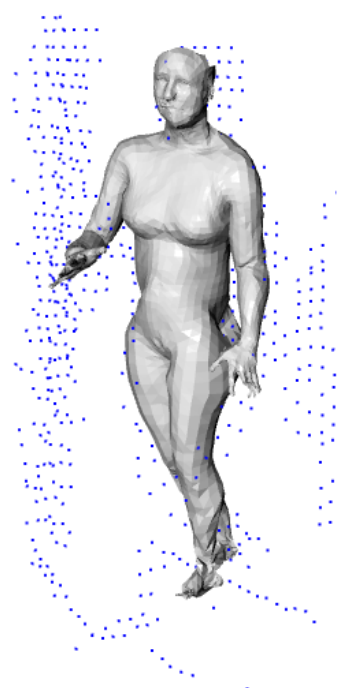}} & \adjustbox{valign=c}{\includegraphics[width=0.09\linewidth]{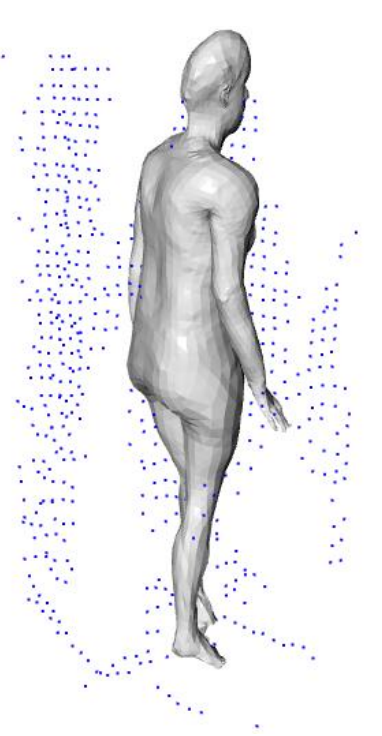}} & \adjustbox{valign=c}{\includegraphics[width=0.09\linewidth]{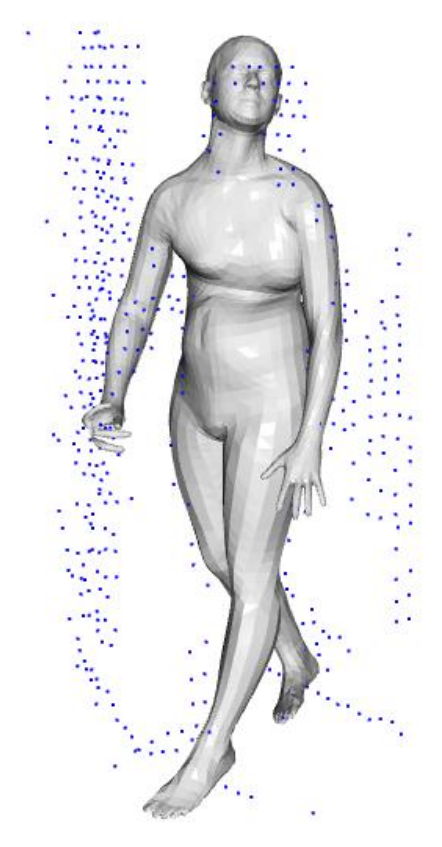}} & \adjustbox{valign=c}{\includegraphics[width=0.09\linewidth]{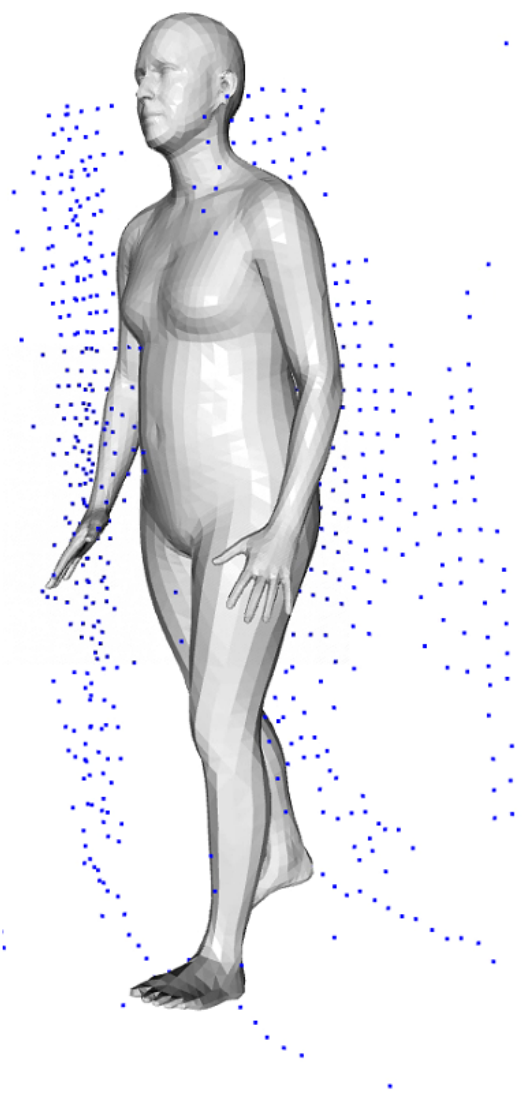}} & \adjustbox{valign=c}{\includegraphics[width=0.09\linewidth]{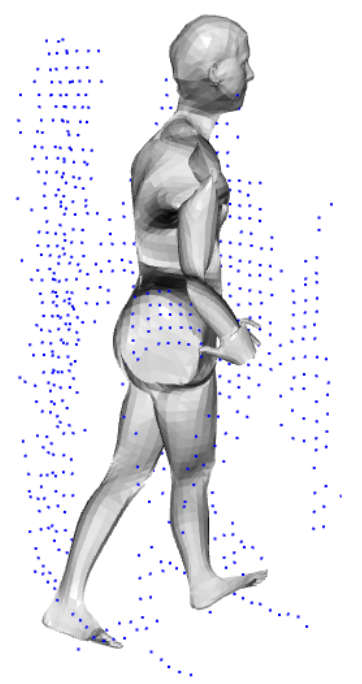}}
\\
\adjustbox{valign=c}{\includegraphics[width=0.10\linewidth]{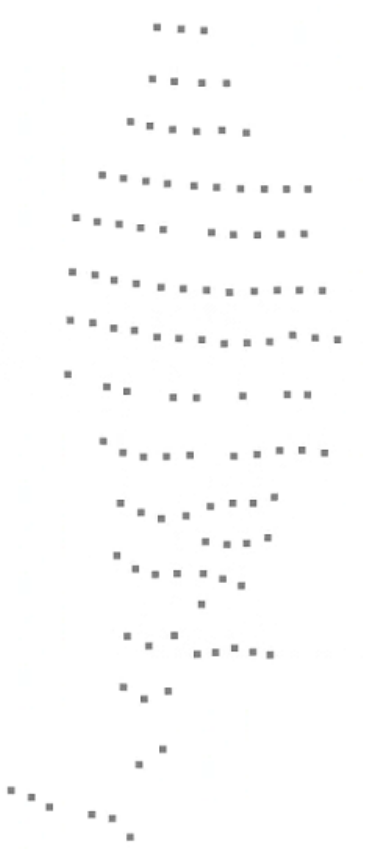}} & \adjustbox{valign=c}{\includegraphics[width=0.09\linewidth]{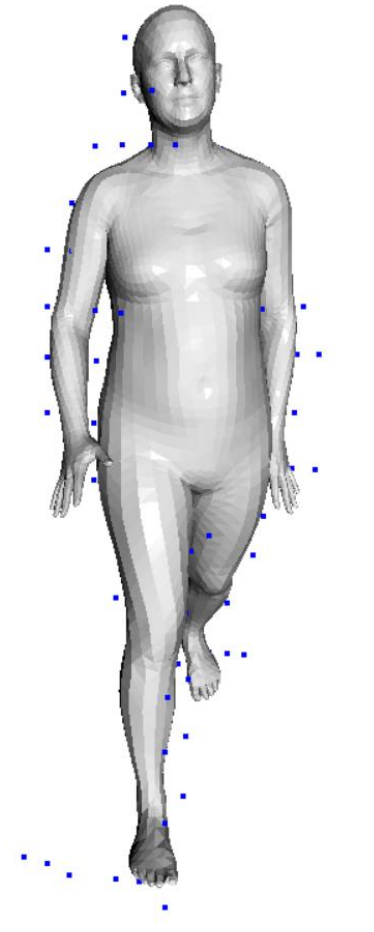}} & \adjustbox{valign=c}{\includegraphics[width=0.12\linewidth]{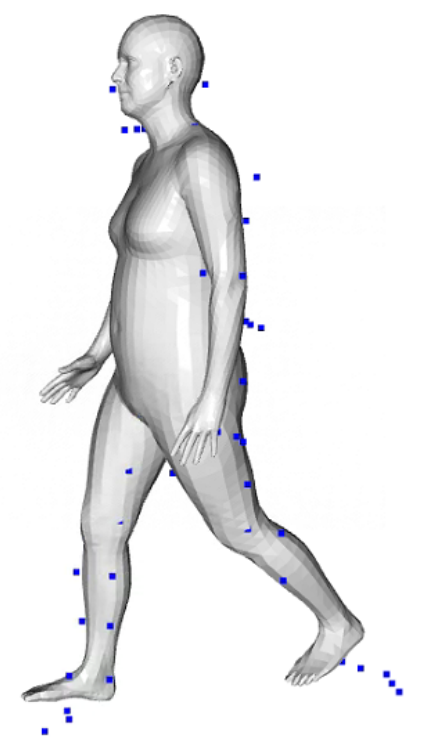}} & \adjustbox{valign=c}{\includegraphics[width=0.085\linewidth]{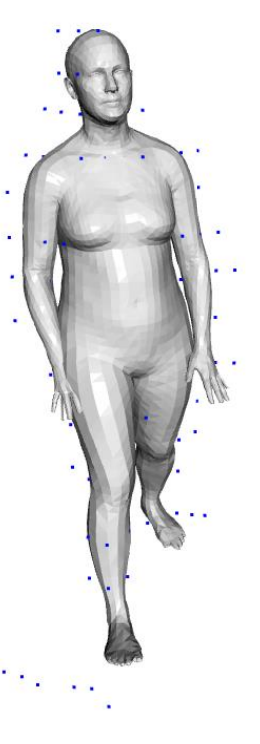}} & \adjustbox{valign=c}{\includegraphics[width=0.09\linewidth]{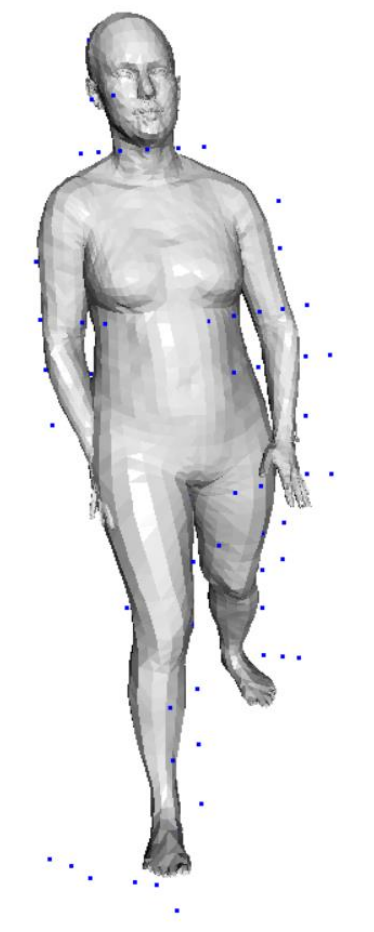}} & \adjustbox{valign=c}{\includegraphics[width=0.082\linewidth]{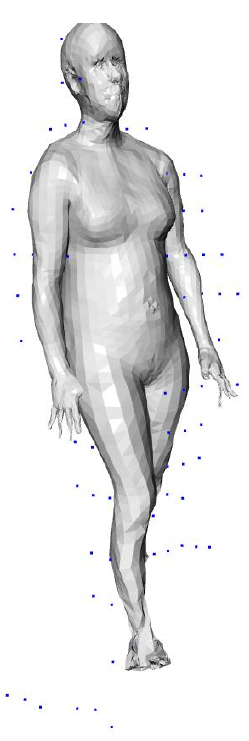}} & \adjustbox{valign=c}{\includegraphics[width=0.09\linewidth]{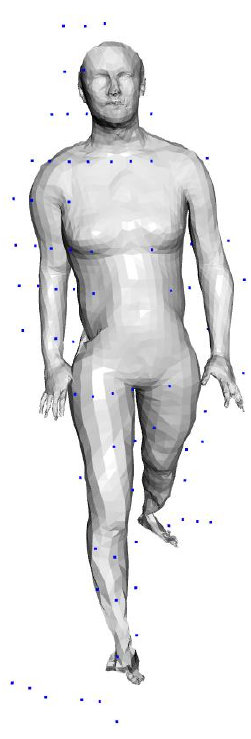}} & \adjustbox{valign=c}{\includegraphics[width=0.085\linewidth]{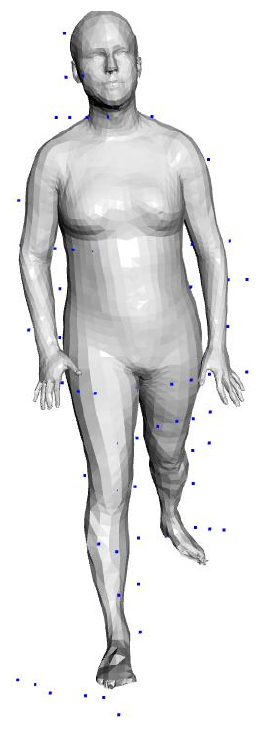}} & \adjustbox{valign=c}{\includegraphics[width=0.09\linewidth]{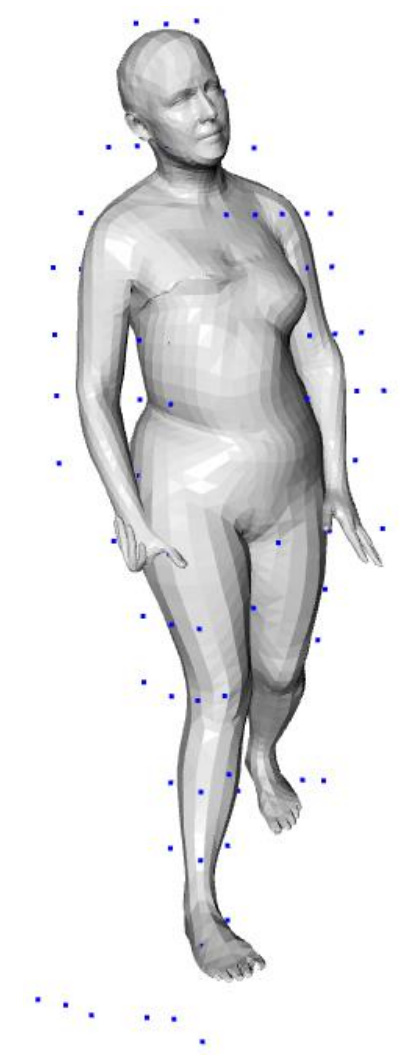}} & \adjustbox{valign=c}{\includegraphics[width=0.09\linewidth]{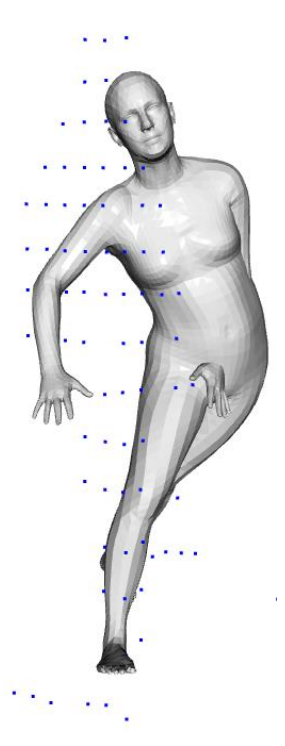}} & \adjustbox{valign=c}{\includegraphics[width=0.085\linewidth]{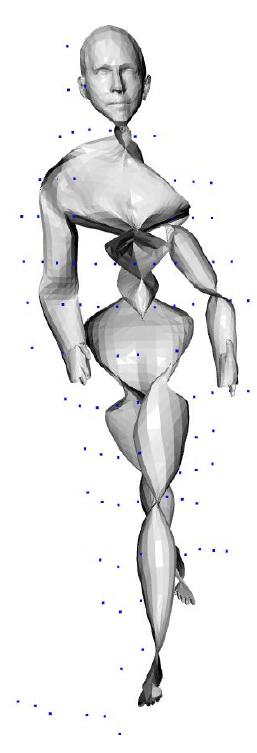}}
\\
{\includegraphics[width=0.09\linewidth]{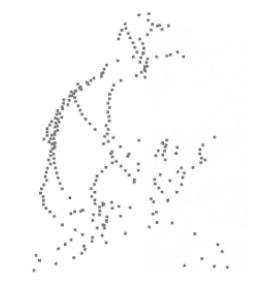}} & {\includegraphics[width=0.09\linewidth]{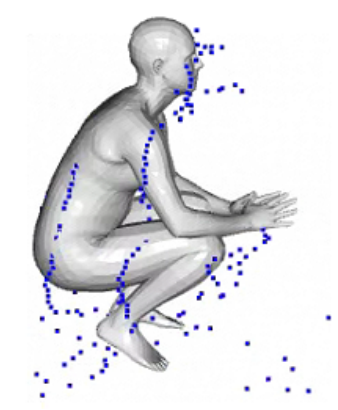}} & {\includegraphics[width=0.10\linewidth]{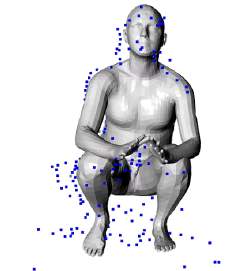}} & {\includegraphics[width=0.09\linewidth]{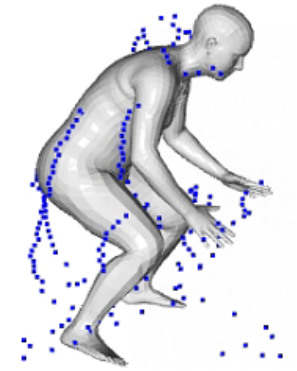}} & {\includegraphics[width=0.095\linewidth]{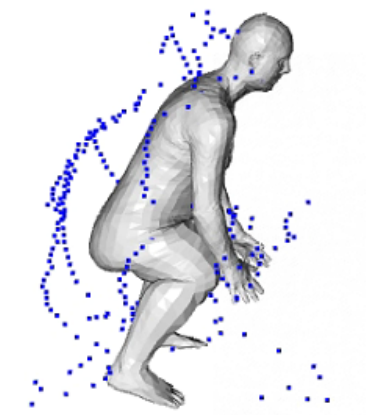}} &{\includegraphics[width=0.09\linewidth]{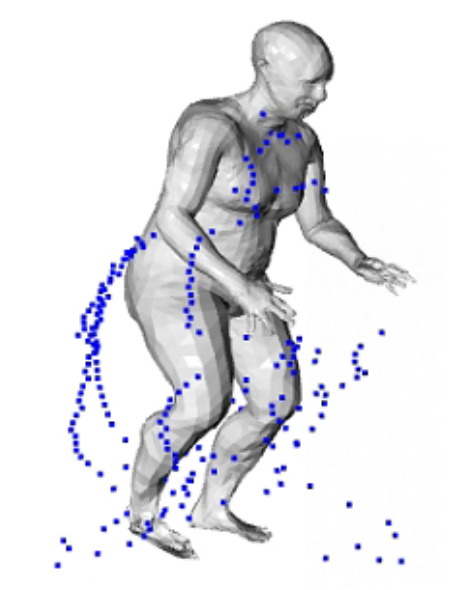}} & {\includegraphics[width=0.08\linewidth]{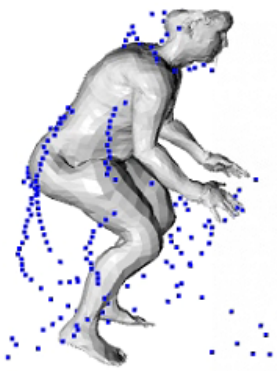}} & {\includegraphics[width=0.09\linewidth]{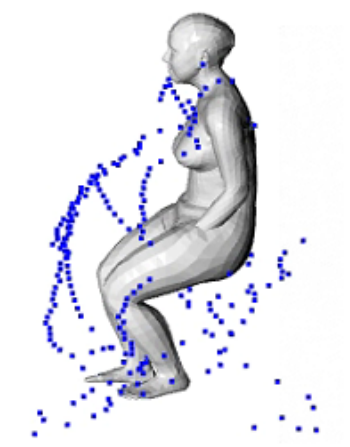}} & {\includegraphics[width=0.09\linewidth]{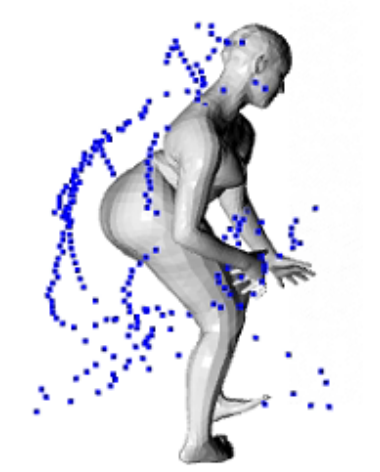}} &{\includegraphics[width=0.09\linewidth]{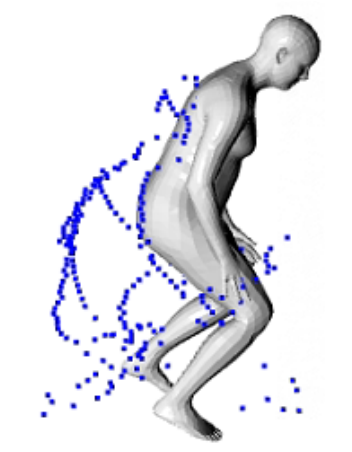}} & {\includegraphics[width=0.09\linewidth]{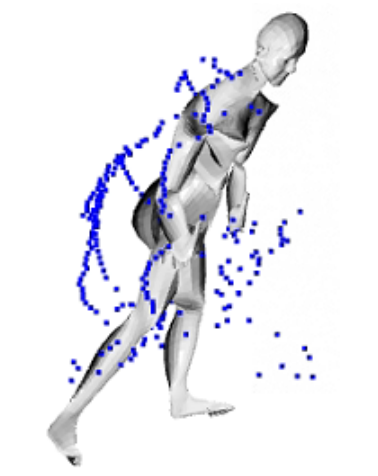}} 
\\
{\includegraphics[width=0.09\linewidth]{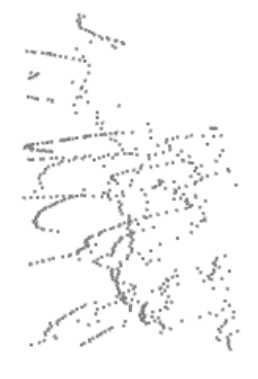}} & {\includegraphics[width=0.09\linewidth]{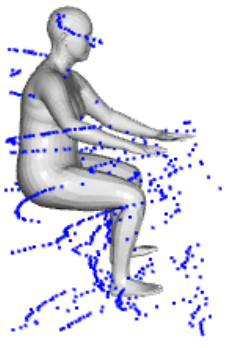}} & {\includegraphics[width=0.09\linewidth]{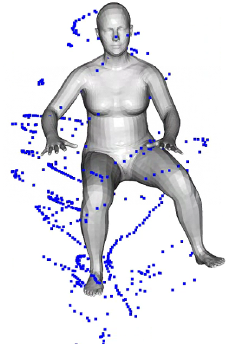}} & {\includegraphics[width=0.09\linewidth]{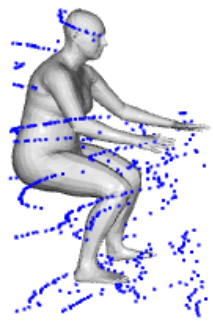}} & {\includegraphics[width=0.09\linewidth]{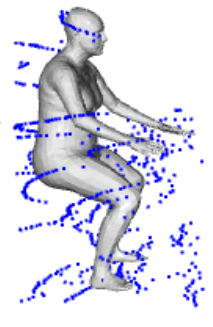}} &{\includegraphics[width=0.085\linewidth]{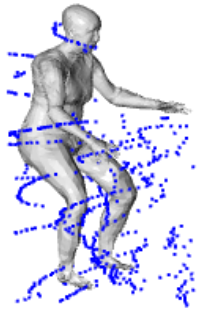}} & {\includegraphics[width=0.09\linewidth]{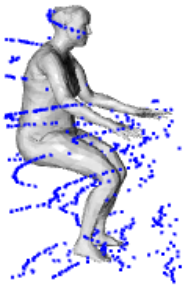}} & {\includegraphics[width=0.09\linewidth]{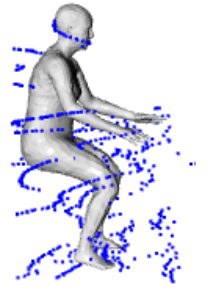}} & {\includegraphics[width=0.09\linewidth]{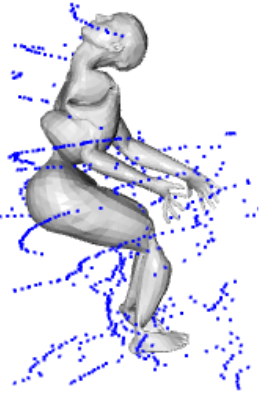}} &{\includegraphics[width=0.09\linewidth]{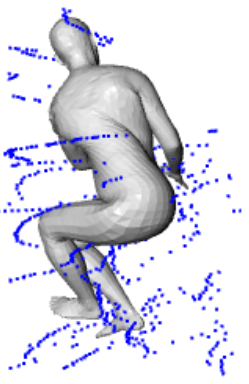}} & {\includegraphics[width=0.09\linewidth]{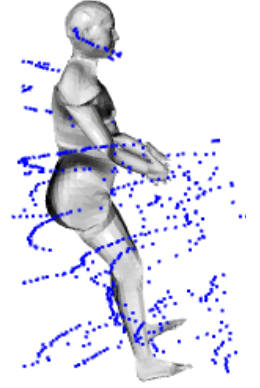}} 
\end{tabular}
}
\centering
% \vspace{-6mm}
\caption{Comparisons with the state-of-the-art methods on the Waymo, SLOPER4D and Human-M3 datasets. The meshes are visualized with the input point clouds (blue points) for better comparison. The top three rows show examples corresponding to challenges of incomplete point clouds (low leg missing), noise, and sparse point clouds. The bottom two rows show more diverse human poses that exist in the Human-M3 dataset. (GT: ground-truth, {GT\_V: ground-truth (alternative view)}, LMR: LiDAR-HMR (PRN+MRN+MeshIK), PM: PRN+MRN, P2M: Pose2Mesh \cite{choi2020pose2mesh}, PP: PRN+P2M, and PPM: PRN+P2M+MeshIK).}
% \vspace{-8mm}
\label{fig::diff_compare}
\end{figure*}

For \textbf{HMR}, the qualitative comparison and quantitative evaluation results are illustrated in Fig. \ref{fig::diff_compare} and Table \ref{table:table_sota}, respectively. 
Both HybrIK and LiDARCap reconstruct the SMPL pose parameters on the basis of the skeleton, but they suffer from inaccurate local details, which are generally reflected by the unfaithful axial rotations of the bones under the correct skeleton. This is mainly because they fail to extract precise local details from the point clouds. VoteHMR and SAHSR do not achieve satisfactory results. Both methods directly regress the human body mesh from the point clouds without sparse-to-dense modeling. They are parameterized methods that face difficulties in effectively modeling the relationships between point clouds and corresponding human meshes. For the more challenging datasets such as Waymo and Human-M3, larger errors result in unsatisfactory MPERE scores. Comparing the above two methods highlights the effectiveness of the proposed point cloud-to-SMPL reconstruction pipeline. Specifically, LiDAR-HMR also performs well in reconstructing more complex and diverse poses, demonstrating adequate robustness and portability. The final throughput of the network is approximately 7.5 frames per second (fps), which is close to that of Pose2Mesh+MeshIK. In particular, when the MeshIK network structure is excluded, the network's throughput capability is 12.35 fps, which is slightly lower than that of Pose2Mesh's 14.38 fps. This can be attributed mainly to the deeper and more complex feature processing introduced by the MRN. However, it is important to note that the typical data frame rate of commonly used LiDAR systems is approximately 10 fps. Therefore, the network's throughput capability can meet the application requirements.

For Pose2mesh, although both methods are based on a similar pipeline of sparse-to-dense modeling, the MRN takes the point cloud as input to assist in the reconstruction. With the same backbone and training conditions, the MRN outperforms the MeshNet proposed in Pose2Mesh while consuming fewer computational resources.  Additionally, constraints are applied to the intermediate reconstruction results, making the entire reconstruction process more rational and effective. As illustrated in Fig. \ref{fig::diff_compare}, what cannot be quantitatively measured from the metrics is that Pose2Mesh often suffers from unreasonable body local details. Although the Pose2Mesh results on the LiDARHuman26M dataset are higher in MPJPE and MPVPE than for LiDAR-HMR, their higher MPERE also indicates a common irrational situation of regressed human poses. This is mainly because the connection relationship of vertex points is not considered in the upsampling process, which makes it difficult for the network to model the semantic information of each vertex point.

Finally, the meaning of \textbf{MeshIK} is clear, and the rationality and accuracy of the estimated poses are improved after MeshIK's processing. This is reflected in MPERE's improvement and more reasonable visualization of the results. After MeshIK, the original vertex coordinate output of the model is transformed into semantically SMPL shape and pose parameters, which describe the posture of the human body. Compared with the original representation, this representation can be used for other tasks such as timing processing or motion recognition. In general, the proposed LiDAR-HMR achieves the best results with fewer computing resources under four datasets and different settings.

\subsection{Ablation Study}
We conduct an ablation study on the Waymo dataset \cite{sun2020scalability}.

\textbf{Ablation of the PRN.}
We compare the following different settings:
(1) Without the vote module, the output features of PointTransformer-v2 are fed into a fully connected layer to directly output the human pose. (called ``no\_vote'')
(2) The refinement module is removed. (called ``no\_refine'')
(3) With the condition of ``no\_vote'', the refinement module is removed. (called ``no\_all'')
(4) The loss terms of $L_\mu$ and $L_q$ are removed from $L_{PRN}$ in equation \ref{eq_prn} (called ``no\_voteloss'').

The quantitative evaluation results are listed in Table \ref{table:table_ablation_prn}. Compared with the ``no\_all'' setting, the introduction of the vote module results in an MPJPE improvement of 3.09 cm and a PA-MPJPE improvement of 1.69 cm. Similarly, adding the refinement module can also result in improvements of 3.25 cm and 2.24 cm respectively. Notably, both the refinement module and the vote module can achieve good results when used alone. They obtain better human poses from the two aspects of ``more accurate estimation'' and ``more accurate adjustment''. Even so, the introduction of the vote module can still reduce the computational consumption of the network. The vote loss term in equation \ref{eq_prn} has a slight benefit on the performance of the PRN, and it is more natural and direct for the supervised backbone to learn point-by-point features.

PA-MPJPE measures only the relative error of human skeletons, reflecting the network's ability to perceive overall human poses. Intuitively, under the ``no\_vote'' setting, the network lacks access to more local human detail, thus focusing more on reconstructing the relative 3D skeleton as a whole; this leads to improved performance in PA-MPJPE. However, it is important to note that PA-MPJPE does not account for errors in the location of the body in global coordinates. In practical applications, especially in the subsequent human mesh reconstruction process, obtaining accurate 3D skeletons in global coordinates is highly important.

\begin{figure*}[t]
 % \vspace{-4mm}
 \subfloat[MPVPE]{\includegraphics[width=0.5\textwidth]{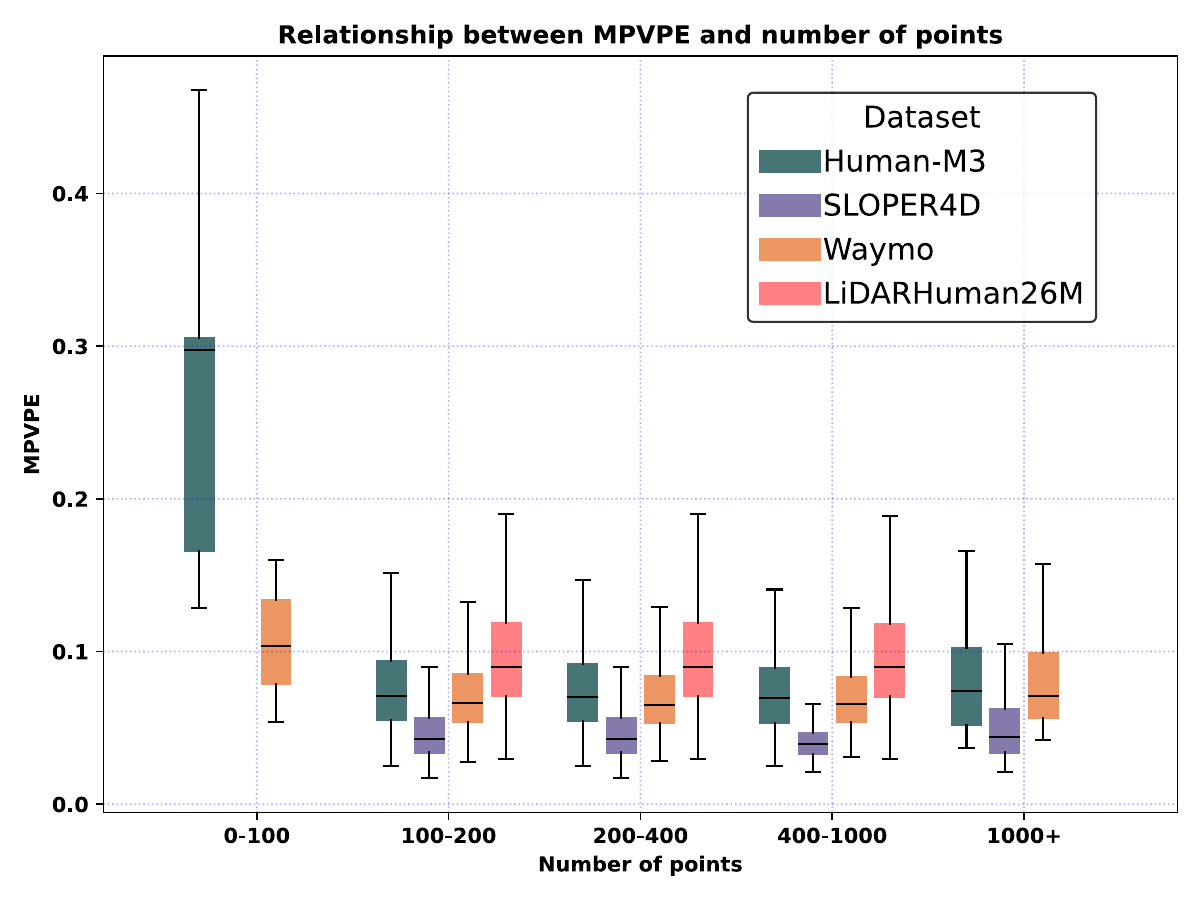}
  \label{fig:point_MPVPE}}
\subfloat[MPERE]{\includegraphics[width=0.5\textwidth]{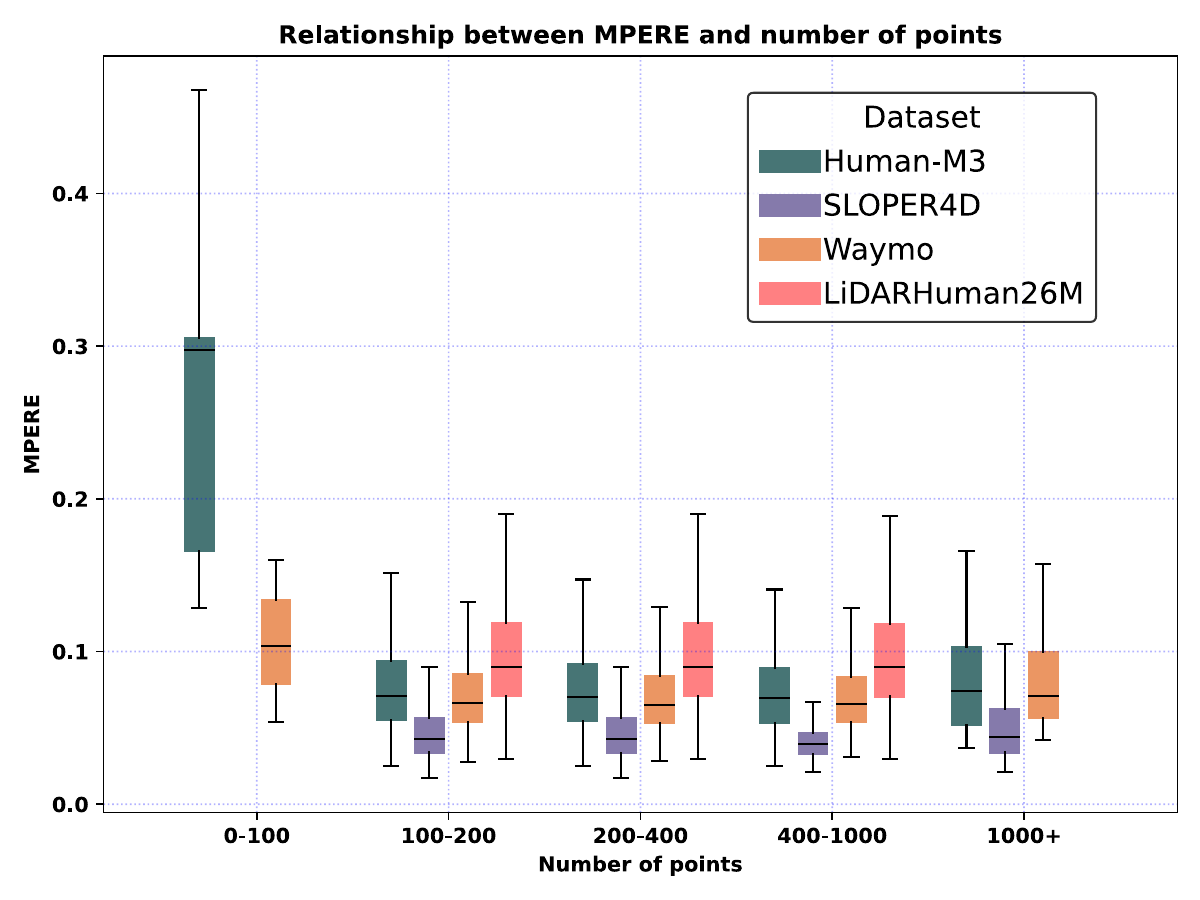}
  \label{fig:point_MPERE}}
    \caption{Analysis of the relationship between performance and the number of points.}
\vspace{-4mm}
\label{fig:point_metric}
\end{figure*}

\begin{table}[t]
\setlength{\tabcolsep}{3.0pt}
\caption{Ablation evaluation of the PRN on the Waymo dataset. The metrics include MPJPE (cm), PA-MPJPE (cm), and the computation costs while training when the batch size is set to 1 (GFLOPs). The best values are shown in bold.}
% \vspace{-4mm}
\begin{center}
% \resizebox{0.48\textwidth}{!}
{
\begin{tabular}{ c c  c  c} 
\toprule
 Group & MPJPE & PA-MPJPE & GFLOPs \\ 
\midrule
no\_vote & 6.89 & \textbf{5.07} & 0.751\\
no\_refine & 7.05 & 5.62 & 0.475\\
no\_all & 10.14 & 7.31 & \textbf{0.347}\\
no\_voteloss & 6.94 & 5.09 & 0.672\\
PRN & \textbf{6.78} & 5.20 & 0.672\\
\bottomrule
\end{tabular}
}
\end{center}
\label{table:table_ablation_prn}
\end{table}

\begin{table}[t]
\setlength{\tabcolsep}{3.0pt}
\caption{Ablation evaluation of the MRN on the Waymo dataset. The metrics include MPJPE (cm), MPVPE (cm), MPERE, and the computation cost while training when the batch size is set to 1 (GFLOPs). The best values are shown in bold.}
\vspace{-4mm}
\begin{center}
% \resizebox{0.48\textwidth}{!}
{
\begin{tabular}{c c c  c  c  c } 
\toprule
 & Group & MPJPE (PRN) & MPVPE & MPERE & GFLOPs \\ 
\midrule
\multirow{3}{*}{PRN Features} & no\_PRN & - & 28.03 & 0.306 & 2.225 \\ 
 & no\_pretrain & 6.62 & 8.53 & 0.128 & 2.225 \\
 & no\_pcd & 7.05 & 9.05 & 0.145 & 2.103\\
 \midrule
\multirow{3}{*}{P2M Structures} & no\_MRN & 6.86 & 10.22 & \textbf{0.057} & \textbf{0.750} \\ 
& graphormer & 8.05 & 9.83 & 1.760 & 9.150 \\
& upsample & 6.49 & 8.34 & 0.127 & 2.164 \\
\midrule
\multirow{2}{*}{MRN Losses} & no\_mid & 6.53 & 8.89 & 0.123 & 2.212 \\
& edge\_mid & 6.31 & 8.31 & 0.165 & 2.225\\ 
\midrule
& PRN+MRN & \textbf{6.28} & \textbf{8.24} & 0.119 & 2.225 \\ 
\bottomrule
\end{tabular}
}
\end{center}
\label{table:table_ablation_mrn}
\end{table}

\textbf{Ablation of MRN.}
We compare the performance of the MRN in three different aspects: the utilization of raw point cloud features from the PRN, different structures of the pose-to-mesh (P2M) reconstruction process, and different loss functions of the MRN. Specifically, for the utilization of the PRN features we utilize the following settings:
(1) The PRN is initialized with pretrained weights and no longer participates in end-to-end training, and the loss function does not include the pose estimation loss of the PRN (called ``no\_PRN''). (2) The PRN is not pretrained and initialized with random weights (called ``no\_pretrain''). (3) The point cloud features are not fed into the reconstruction process of the MRN, and the self-attention modules at each resolution only calculate the correspondence between vertex features (called ``no\_pcd''). 
For P2M structures, we utilize the following settings:
(1) The output features of the PRN are directly fed into an MLP-based network to regress the SMPL pose and shape parameters (called ``no\_MRN'').
(2) The output features of the PRN are fed into a graphormer-based network consistent with that in \cite{lin2021mesh} to upsample and regress vertex coordinates (called ``graphormer'').
(3) Different resolutions in the MRN network use upsampling (consistent with Pose2Mesh) instead of feature propagation (called ``upsample'').
For losses, we utilized the following settings:
(1) The loss function does not include the mesh reconstruction loss at intermediate resolutions (called ``no\_mid''), and (2) the reconstruction loss at intermediate resolutions not only constrains the corner point positions but also introduces constraints on edge lengths ($L_F$ term in Section \ref{sec:mrn}) (called ``edge\_mid'').

\begin{figure*}[t]
\centering
\subfloat[Human-M3 dataset.]{\includegraphics[width=0.345\textwidth]{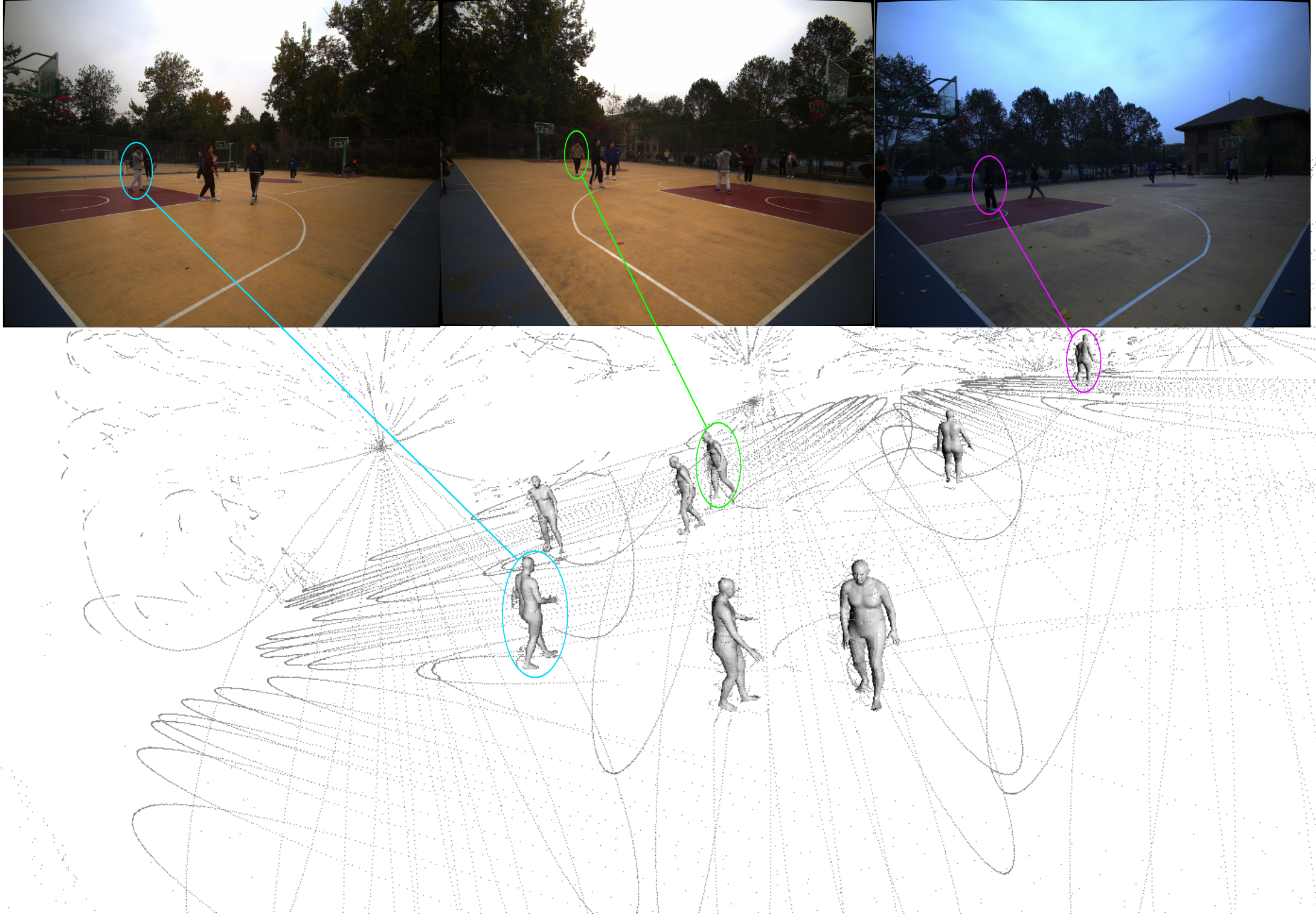}
\label{fig:humanm3}}
\subfloat[Waymo dataset.]{\includegraphics[width=0.332\textwidth]{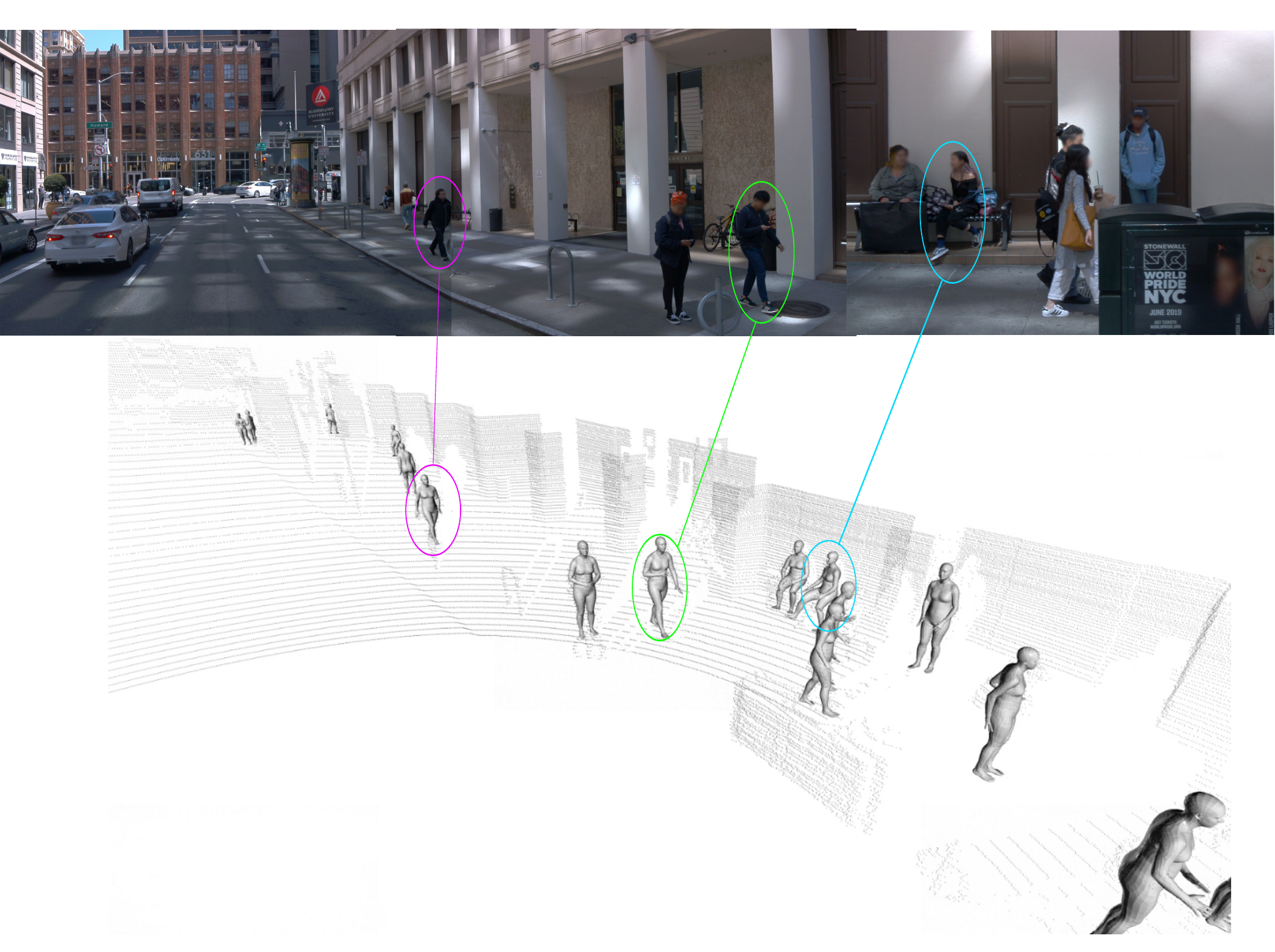}
\label{fig:waymo}}
\subfloat[THU-MultiLiCa dataset.]{\includegraphics[width=0.285\textwidth]{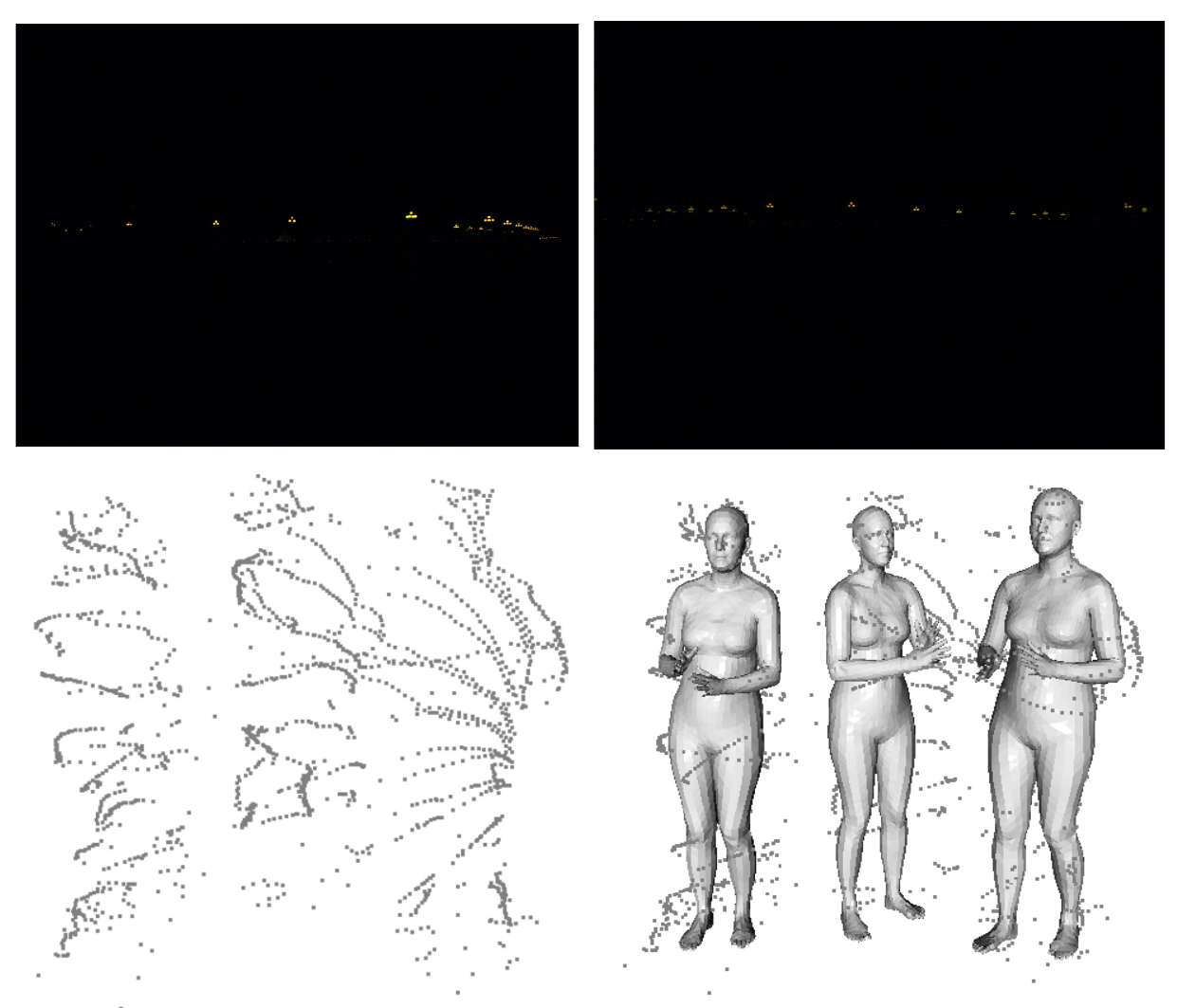}
\label{fig:thu_lica}}
\caption{More visual results of LiDAR-HMR in multiperson scenes. (a) Human-M3 dataset. (b) Waymo dataset. (c) THU-MultiLiCa \cite{zhang2022flexible} dataset (top: RGB images; bottom left: point clouds; bottom right: estimated meshes overlapping with the point clouds).}
\label{fig:more_demo}
\vspace{-4mm}
\end{figure*}

\begin{figure}[t]
 \vspace{-5mm}
        \centering
        \subfloat[Collapsed local details.]{\includegraphics[width=0.22\textwidth]{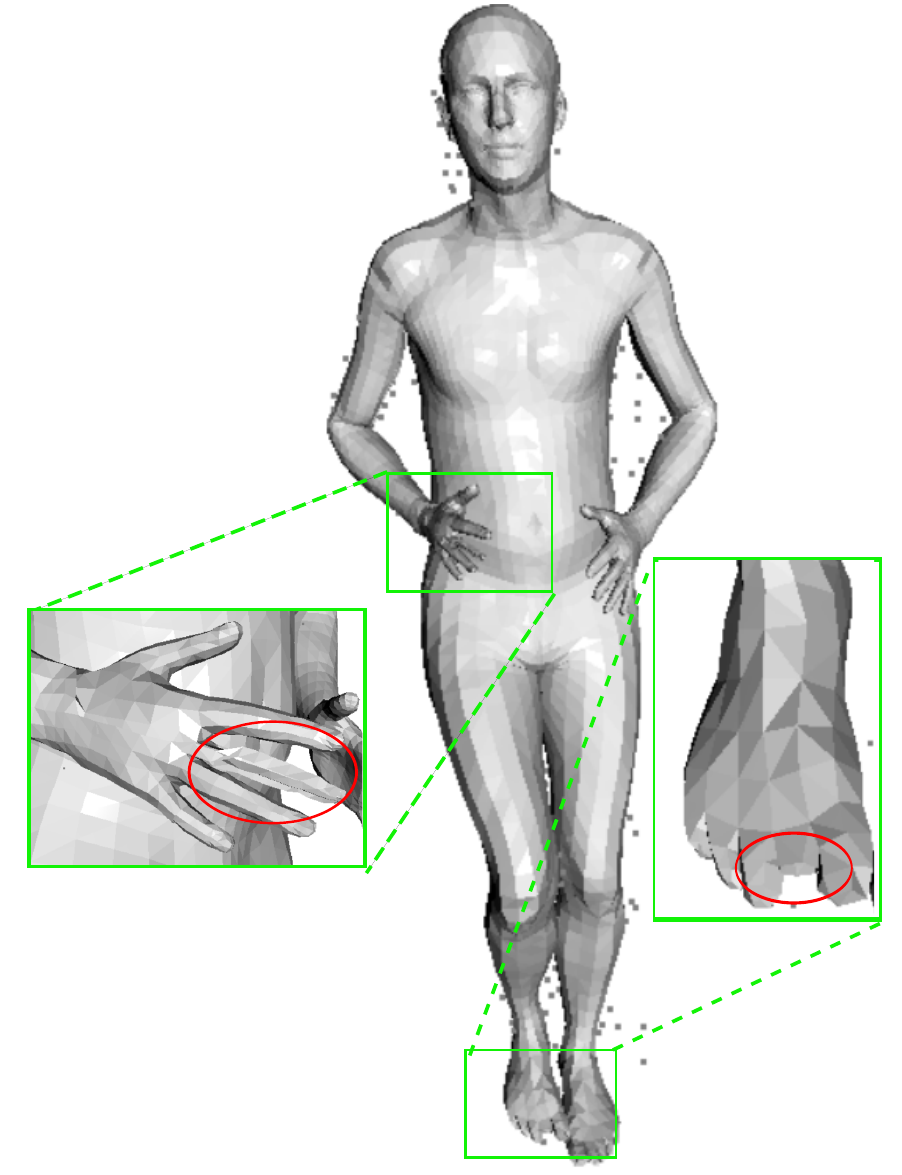}
        \label{fig:sub1}}
        \subfloat[Partial missing.]{\includegraphics[width=0.28\textwidth]{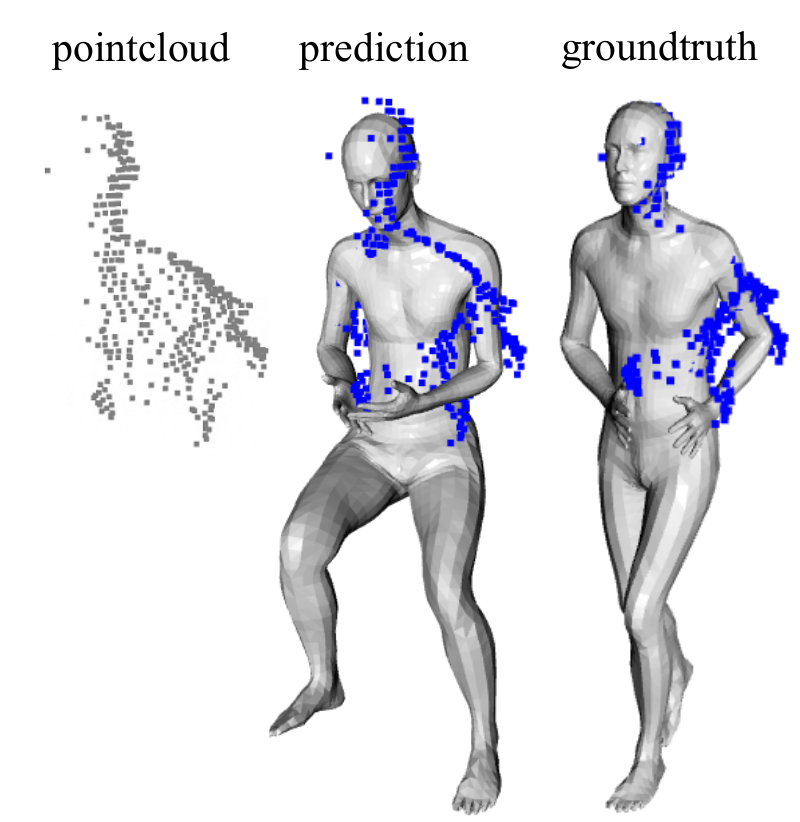}
        \label{fig:wrong_pose}}
        
        \caption{Two kinds of typical failure cases. In (a), red ellipses mark the collapsed details.}
        \label{fig:failure}
    \vspace{-5mm}
\end{figure}

The quantitative evaluation results are illustrated in Table \ref{table:table_ablation_mrn}.
For \textbf{the utilization of raw point cloud features from the PRN}, first, end-to-end training is essential for the MRN. When the backbone does not participate in end-to-end training, the performance of the MRN decreases drastically, which indicates that the features learned from pose estimation pretraining cannot represent 3D mesh reconstruction. Therefore, weight updating is important for the backbone of MRN training. Second, the effect of using raw point cloud features is obvious; when comparing the settings ``PRN+MRN'' and ``no\_pcd'', the MPVPE and MPERE decreased by 0.81 cm and 0.026, respectively, accounting for 8.95\% and 17.93\%. This demonstrates the effectiveness of the proposed strategy of integrating point clouds into the reconstruction process. Finally, 
the pre-training process is also beneficial for mesh reconstruction; the MPVPE and MPERE decreased by 0.29 cm and 0.009, respectively, accounting for 3.39\% and 7.03\%. This finding shows that the pretraining strategy can provide more semantic and prior information for mesh reconstruction, which is helpful.

Regarding the \textbf{pose-to-mesh network structures}:
First, sparse-to-dense modeling is essential. Regressing parameters directly from the PRN features results in unsatisfactory performance. Second, inside sparse-to-dense reconstruction, the proposed approach of feature propagation in different mesh resolutions showed improvements over the upsampling method in Pose2Mesh and Graphormer, with MPVPE and MPERE decreasing by 0.1 cm and 0.008, respectively, compared with those in Pose2Mesh. This shows that the proposed feature propagation module inherits the information between different resolutions more effectively in the process of feature conduction, and achieves better results.

For \textbf{the MRN losses} of intermediate meshes, it was observed that supervising the mesh at intermediate resolutions has a slight effect on the algorithm's performance. This may be attributed to providing the model with a more reasonable reconstruction process at intermediate stages, facilitating better learning. Interestingly, imposing constraints on the mesh length at intermediate resolutions has a mildly negative effect on the model performance, possibly because excessive constraints during the intermediate reconstruction process lead to reduced freedom in the final mesh reconstruction.

\textbf{Analysis of the number of points.}
%The number of input points also affects the performance of LiDAR-HMR. 
As shown in Fig. \ref{fig:point_metric}, the performance of LiDAR-HMR is relatively stable when the number of points exceeds 100. When the number of points is less than 100, the algorithm exhibits significant fluctuations in MPVPE and MPERE on the Human-M3 and Waymo datasets, respectively. The performance of LiDAR-HMR is indeed affected when the input points are sparse, and problems under such circumstances are more challenging. We found that the three public datasets have a limited number of samples with sparse points (less than 100), which may be a contributing factor to this phenomenon as the learning process is not sufficiently comprehensive in handling such cases. This issue can be addressed by extracting challenging samples. 

\subsection{Limitations and Failure Cases}
Failure cases are illustrated in Fig. \ref{fig:failure}. Specifically, although the nonparametric representation can reconstruct the human mesh more accurately and intuitively than the parametric representation can, the mesh vertices regressed by the MRN still suffer from unsatisfactory local details which are collapsed and not smooth in some cases. This makes the reconstructed mesh unrealistic (Fig. \ref{fig:sub1}).

In addition, the incompletion of the input point cloud can lead to errors in the overall mesh. LiDAR-HMR is capable of mitigating the issues arising from missing portions of the point cloud to some extent, particularly through the network's ability to reconstruct the missing parts of the pose on the basis of effective pose priors learned from the data. However, when a certain body part is entirely missing, LiDAR-HMR still struggles to achieve fully accurate results. Specifically, in Fig.\ref{fig:wrong_pose}, the lower half of the body is almost completely absent, which consequently leads to inaccurate estimations via LiDAR-HMR. Such challenging cases are indeed present in real-world applications and are difficult to avoid. Unfortunately, these cases are ambiguous, and introducing additional temporal constraints for pose continuity may be beneficial in addressing these types of issues.

Notably, valid human mesh priors (such as VPoser \cite{pavlakos2019expressive}) are based on parametric representations. Suppose that more human mesh priors can be introduced into network training, or that restrictions can be imposed on the surface mesh of the network output. This approach results in a more reasonable, realistic human surface while ensuring accuracy.

\subsection{More Visual Results}
Apart from Fig. \ref{fig:waymo_demo}, more visualization results of LiDAR-HMR in wide outdoor scenes are illustrated in Fig. \ref{fig:more_demo}. LiDAR-HMR offers robust and satisfactory results on different outdoor scenes, where HMR methods based on RGB images usually struggle. As illustrated in Fig. \ref{fig:thu_lica}, the illumination is too weak to distinguish human bodies in the night scene of the THU-MultiLiCa \cite{zhang2022flexible} dataset, but LiDAR-HMR is not affected. Specifically, this dataset does not provide human pose ground-truth values, so we use the trained weights in the Waymo-v2 dataset for inference, which also shows the robustness and portability of LiDAR-HMR in different scenes.

\section{Conclusion}
We present the first attempt at 3D human pose estimation and mesh recovery on the basis of single-frame sparse LiDAR point clouds. We solve this problem by proposing a sparse-to-dense reconstruction pipeline. We introduce the input point cloud to assist the whole reconstruction process, which imposes constraints on the intermediate results of the reconstruction and achieves good results. In addition, we propose MeshIK to regress the SMPL parameters from unconstrained mesh vertices, which not only improves the rationality of the meshes but also imparts semantics to the output mesh. However, the proposed method still fails in some very sparse scenarios and can output unreasonable self-interspersed meshes. A possible solution is to consider more human poses a priori during the reconstruction process. We hope that our work will inspire subsequent work on sparse point clouds or multimodal human perception tasks.

\section*{Acknowledgments}
This work was supported by the National Natural Science Foundation of China under Grant 62321005 and the National Key Research and Development Program of China under Grant (2018AAA0102803).

\bibliographystyle{splncs04}
\bibliography{main}

\end{document}